\documentclass[letterpaper]{article}
\usepackage[preprint]{aaai2027}
\usepackage{rotating}
\usepackage{pdflscape}
\usepackage{graphicx}
\usepackage{booktabs}
\usepackage{multirow}
\usepackage{colortbl}
\usepackage{array}
\usepackage{amsmath}
\usepackage{microtype}
\usepackage[numbers,sort&compress]{natbib}
\usepackage{caption}
\usepackage{subcaption}
\usepackage{enumitem}

\definecolor{lightblue}{RGB}{213,232,240}
\definecolor{lightyellow}{RGB}{255,243,205}
\definecolor{lightred}{RGB}{255,224,224}
\definecolor{findingblue}{RGB}{0,70,127}
\definecolor{lightgreen}{RGB}{198,239,206}

\setlist{nosep,leftmargin=*,topsep=1pt,itemsep=0pt}
\newcommand{\finding}[2]{%
  \vspace{2pt}%
  \noindent\textit{\textbf{\textcolor{findingblue}{#1}} \textcolor{findingblue}{#2}}%
  \vspace{2pt}%
}

\title{Selecting Open-Weight Language Models for Zero-Shot Intent Classification:\\A Systematic Evaluation of 41 Models}
\author{Parishruthi Ganesh, Gerry Dozier, Cheryl Seals}
\affiliations{Department of Computer Science and Software Engineering\\Auburn University\\Auburn, Alabama, USA}

\begin{document}

\maketitle

\begin{abstract}
Intent classification is a core component
of task-oriented dialogue systems, yet practitioners have limited
systematic guidance for selecting deployable open-weight language
models under compute, latency, and robustness constraints. We present
a systematic zero-shot evaluation of 41 open-weight language models
spanning 15 families and the 135M--9B parameter range across eight
English single-label intent-classification datasets (a ninth,
ATIS, uses five labeled demonstrations and is reported as an
auxiliary five-shot result), including
standard benchmarks, a large-scale voice-assistant corpus, and
production-derived e-commerce datasets. Beyond exact-match accuracy,
we analyze confidence calibration, robustness to realistic
input perturbations, statistical reliability of model rankings,
deployment efficiency, and benchmark saturation. Our results show
that instruction-tuned 3B models can outperform several evaluated 7B
base models, that differences among leading models on MASSIVE
are statistically indistinguishable under pairwise McNemar tests, and that widely
used benchmarks such as SNIPS have become saturated and no longer
meaningfully discriminate among current open-weight models.
Instruction tuning's effect on confidence calibration is inconsistent
rather than uniformly harmful. These findings provide practical
guidance for selecting and evaluating open-weight language models for
intent classification.
\end{abstract}

\section{Introduction}

Intent classification (IC)---the task of mapping a natural language
utterance to a predefined functional category---is a central
component of task-oriented dialogue
systems~\cite{larson2019clinc,casanueva2020banking77}. When a user says
``Set an alarm for 7am,'' the system must correctly map this to
\texttt{alarm\_set} before any downstream response can be generated.
This mapping underpins virtually every voice assistant, customer support
agent, and conversational AI system deployed at scale.

As large language models (LLMs) have matured, zero-shot IC has become
increasingly attractive~\cite{brown2020gpt3,wei2022flan}: a single
general-purpose model, prompted with a label set and a user utterance,
can perform IC without any task-specific training data. This raises a
practically important question that has received surprisingly little
systematic attention: among the dozens of open-weight LLMs available
in the 135M--9B parameter range that practitioners actually deploy,
\emph{which should they choose, and why?}

The state of the art offers limited guidance here. IntentGrasp~\cite{yin2026intentgrasp},
the most comprehensive recent evaluation, reformulates 49 intent-related
datasets as multiple-choice QA and evaluates both proprietary frontier
models (GPT-5, Gemini-3, Claude-4) and open-weight models---a setting far removed from the compute-constrained,
latency-sensitive environments where most real deployment decisions occur.
To our knowledge, prior work has not systematically evaluated the
open-weight sub-9B space
for deployment-oriented, single-label zero-shot IC.

We address this gap directly. We evaluate \textbf{41 open-weight LLMs}
across \textbf{15 families} (135M--9B parameters) on \textbf{8 zero-shot intent
classification datasets} (plus ATIS as a 5-shot auxiliary result; see
the Overall Model Rankings subsection) spanning academic benchmarks, large-scale
voice-assistant corpora, and real e-commerce deployment data. Our evaluation
goes beyond accuracy to confidence
calibration (ECE, Brier Score), robustness to input noise,
statistical significance of rankings, and
deployment efficiency, all zero-shot with no task-specific
fine-tuning (ATIS excepted). All nine datasets we run follow the single-label classification
paradigm standard in task-oriented dialogue; multi-label intent
detection~\cite{qin2020agif} is outside our scope.

\noindent\textbf{Contributions:} (1) The first systematic zero-shot
evaluation of 41 open-weight LLMs (15 families, 135M--9B) on 8 zero-shot IC
datasets. (2) Confidence intervals and McNemar tests on MASSIVE
showing that the observed differences among the top five models are
not statistically significant. (3) A calibration
analysis for open-weight LLMs on IC, scoped to MASSIVE's compact label
space under free-text generation, showing instruction tuning's effect
on calibration is inconsistent rather than uniformly harmful. (4) Robustness
analysis showing Qwen2.5-7B-Instruct has the lowest observed
typo-induced degradation under the evaluated perturbation seed.
(5) Benchmark saturation analysis recommending de-emphasis of SNIPS
in IC evaluation suites.

\section{Related Work}

Intent classification benchmarks span three decades, from
ATIS~\cite{hemphill1990atis} and SNIPS~\cite{coucke2018snips} to
large-scale corpora such as CLINC150~\cite{larson2019clinc},
Banking77~\cite{casanueva2020banking77}, and
MASSIVE~\cite{fitzgerald2022massive}, with HINT3~\cite{mehta2020hint3}
introducing production e-commerce datasets to expose the gap between
academic benchmarks and deployment reality; Larson and
Leach~\cite{larson2022survey} surveyed this space and noted growing
saturation concerns that our analysis confirms and extends. Zero-shot
classification has improved with model scale and instruction
tuning~\cite{brown2020gpt3,wei2022flan,schick2021exploiting}, with
Mehri and Eric~\cite{mehri2021zeroshot} showing pre-trained LMs can
perform zero-shot intent detection directly. Closest to our setting,
Park et al.~\cite{park2024dynamic} evaluate similarly-scaled
Qwen2.5-1.5B/7B-Instruct and Llama3-8B-Instruct on the same HINT3
datasets we use, but in a 10-shot regime with LLM-driven label
refinement rather than true zero-shot; their finding that Powerplay11
benefits most from label clarification is consistent with our own
finding that Powerplay11 is the hardest dataset in our
zero-shot suite (lowest average and best-model accuracy; see the Benchmark Saturation Analysis). The most comprehensive
recent benchmark, IntentGrasp~\cite{yin2026intentgrasp}, evaluates 20
frontier LLMs on 49 reformatted datasets as multiple-choice QA;
ConsintBench~\cite{li2025consint} and SessionIntentBench~\cite{yang2025sessionintent}
address e-commerce and session-level intent, but their underlying
datasets are not publicly
available. Our
work is complementary, focusing on open-weight sub-9B models with
deployment-oriented analyses (calibration, robustness, efficiency,
ranking reliability) absent from prior benchmarks. Calibration
failures in modern neural networks are well
documented~\cite{guo2017calibration,zhao2021calibrate}, but
instruction tuning's effect on calibration has received limited
attention for open-weight LLMs on IC specifically, which we study
directly. An extended discussion of related work, including
additional citations and closer comparisons to each prior benchmark,
appears in the Supplementary Material.

\section{Datasets}

We evaluate on nine single-label intent classification datasets selected
to span a wide range of domains, label space sizes, and collection
settings; eight of these are used zero-shot in all aggregate scores,
rankings, and comparisons in this paper, and the ninth (ATIS) is
reported separately as a 5-shot auxiliary result (Overall Model Rankings subsection). SGD was excluded because its single-turn inputs lack the
dialog context required for reliable intent signal.

Table~\ref{tab:datasets} presents the full dataset statistics. The
suite includes four standard academic benchmarks (CLINC150, Banking77,
ATIS, SNIPS; ATIS is 5-shot, the other three zero-shot), one large-scale voice-assistant production corpus (MASSIVE),
one hierarchical parsing benchmark (MTOP), and three real e-commerce
production datasets from the HINT3 corpus~\cite{mehta2020hint3} that
represent genuine deployment conditions. For each dataset we evaluate
on the first 500 test examples, selected deterministically by index
(Supplementary Material), with one exception: the three HINT3
datasets use their complete test sets (459, 309, and 253 examples for
Curekart, Powerplay11, and Sofmattress, respectively).

\begin{table*}[t]
\centering
\caption{Dataset statistics. Avg Acc = mean accuracy across the
evaluated model cohort under each dataset's reported prompting
protocol (Datasets section); all 41 models have completed
MASSIVE inference. ATIS$^{\dagger}$ is evaluated five-shot and is
excluded from all zero-shot aggregates and rankings; the other
eight datasets are zero-shot. Len = mean utterance word count. All datasets follow the single-label
classification paradigm.}
\label{tab:datasets}
\setlength{\tabcolsep}{6pt}
\begin{tabular}{llrrrl}
\toprule
\textbf{Dataset} & \textbf{Domain} & \textbf{\#Intents} & \textbf{Len (words)} & \textbf{Avg Acc} & \textbf{Source} \\
\midrule
CLINC150 & General assistant (10 domains) & 150 & 8.3 & 0.468 & HuggingFace: \texttt{clinc\_oos} \\
Banking77 & Retail banking & 77 & 11.9 & 0.388 & HuggingFace: \texttt{PolyAI/banking77} \\
ATIS$^{\dagger}$ & Airline travel & 20 & 10.3 & 0.718 & Local CSV \\
SNIPS & Smart home / assistant & 7 & 9.2 & 0.807 & Local CSV \\
MASSIVE & Voice assistant (18 domains) & 60 & 6.8 & 0.432 & \texttt{AmazonScience/massive} \\
MTOP & Facebook Assistant & 117 & 8.5 & 0.348 & Local TSV \\
Curekart & Health supplement e-commerce & 20 & 6.9 & 0.364 & HINT3 (Local CSV) \\
Powerplay11 & Fantasy cricket gaming & 53 & 7.0 & 0.299 & HINT3 (Local CSV) \\
Sofmattress & Mattress e-commerce & 20 & 6.6 & 0.378 & HINT3 (Local CSV) \\
\bottomrule
\end{tabular}
\end{table*}

\section{Evaluation Framework}

\paragraph{Models.} We evaluate 41 open-weight LLMs across 15 model
families spanning 135M to 9B parameters, one family per distinct
model-developer/architecture lineage: Qwen2/2.5~\cite{qwen2_2024,qwen25_2024},
Llama 3/3.1/3.2~\cite{llama3_2024}, Mistral~\cite{mistral7b_2023},
Yi~\cite{yi_2024}, Phi~\cite{phi3_2024}, InternLM2~\cite{internlm2_2024},
DeepSeek-R1-distilled~\cite{deepseekr1_2025}, ChatGLM3-6B~\cite{chatglm_2024},
MiniCPM~\cite{minicpm_2024}, SmolLM2~\cite{smollm2_2025}, Pythia~\cite{pythia_2023}, BLOOM~\cite{bloom_2022},
StableLM~\cite{stablelm2_2024}, TinyLlama~\cite{tinyllama_2024}, and
GPT-J~\cite{gptj_2021}. ChatGLM3-6B
completed evaluation successfully and is included throughout. Model families with
vLLM-incompatible architectures (Gemma-2, Falcon, and OLMo-2, all
verified to fail to load under vLLM~0.6.3 across all evaluated sizes)
were not evaluated. All 41 evaluated models have complete main-evaluation
accuracy scores. Models are categorized as \emph{Base} (pre-trained
only), \emph{Instruct} (instruction-tuned), or \emph{Reasoning}
(chain-of-thought optimized). The complete ranked model list appears in
the Supplementary Material.

\paragraph{Inference Protocol.} All models use vLLM
0.6.3~\cite{kwon2023vllm} on an NVIDIA V100-32GB GPU with float16
precision, temperature 0.0 (greedy decoding), and a maximum of 20
output tokens. All nine datasets, including MASSIVE, use free-text
generation with exact-match parsing after normalization; we do not
apply constrained decoding. MASSIVE's compact label space (60 short,
one-to-two word intents; Table~\ref{tab:datasets}) is nonetheless the
basis for our calibration analysis, discussed next.

\paragraph{Prompt Design.} Eight datasets use zero-shot prompts
consisting of a role description, the complete valid-label list, an
output instruction, and the input utterance. The ATIS prompt
additionally includes five labeled demonstrations and is treated
only as an auxiliary five-shot evaluation. No chain-of-thought
instructions or task-specific parameter tuning are used.
Full prompt templates are given in the Supplementary Material.

\paragraph{Scoring.} The primary metric is exact-match accuracy after
normalization (lowercase, article removal, punctuation stripping).
Secondary metrics include Expected Calibration Error (ECE) and a
binary Brier score for correctness confidence (Confidence Calibration subsection).
The aggregate score is computed as the mean across six dataset groups:
the standard IC composite (mean of CLINC150, Banking77, and SNIPS;
ATIS is excluded from this composite and from the aggregate entirely,
see the Overall Model Rankings subsection),
MASSIVE, MTOP, Curekart, Powerplay11, and Sofmattress. We refer to
this as a \emph{group-balanced} aggregate: the three zero-shot academic
benchmarks are pooled into a single group so that saturated legacy
benchmarks do not outweigh production-oriented datasets; consequently,
each remaining dataset carries the same weight as the entire academic
composite. Per-dataset scores are reported throughout
the Supplementary Material so
alternative aggregations can be computed.

\paragraph{Why Calibration is Restricted to MASSIVE.}
\label{sec:calib_scope}
Calibration requires logprobabilities that represent genuine
label-level confidence, which is only interpretable when generated
tokens correspond to a valid label with reasonable regularity. We use
free-text generation throughout (see Inference Protocol above), so
this is not guaranteed by construction; it is an empirical property we
rely on rather than a decoding-time constraint. MASSIVE's label space
is short and compact (60 intents, predominantly one-to-two word labels
such as \textit{alarm query}), which in practice yields generations
that align closely with the label vocabulary under greedy decoding, so
the sequence-level logprobability of the generated text returned by
the evaluation pipeline is a usable, if
approximate, proxy for the model's confidence in its answer. The other
eight datasets have longer, more heterogeneous label surface forms
(Table~\ref{tab:datasets}), where free-text generations diverge from
the label vocabulary more often and logprobabilities are correspondingly
less interpretable as label-level confidence. Restricting calibration
to MASSIVE is therefore a pragmatic scope decision based on label
compactness, not a claim that MASSIVE's logprobabilities are exact
label-conditional probabilities in the way constrained decoding would
guarantee; we did not apply constrained decoding in this evaluation.
This is a genuine limitation of the calibration analysis, noted again
in the Limitations paragraph below.

\section{Results and Analysis}

\subsection{Overall Model Rankings}

\noindent\textbf{RQ-1:} \textit{Which open-weight models achieve the
highest zero-shot IC accuracy?}

\begin{figure*}[!htb]
\centering
\includegraphics[width=0.85\textwidth]{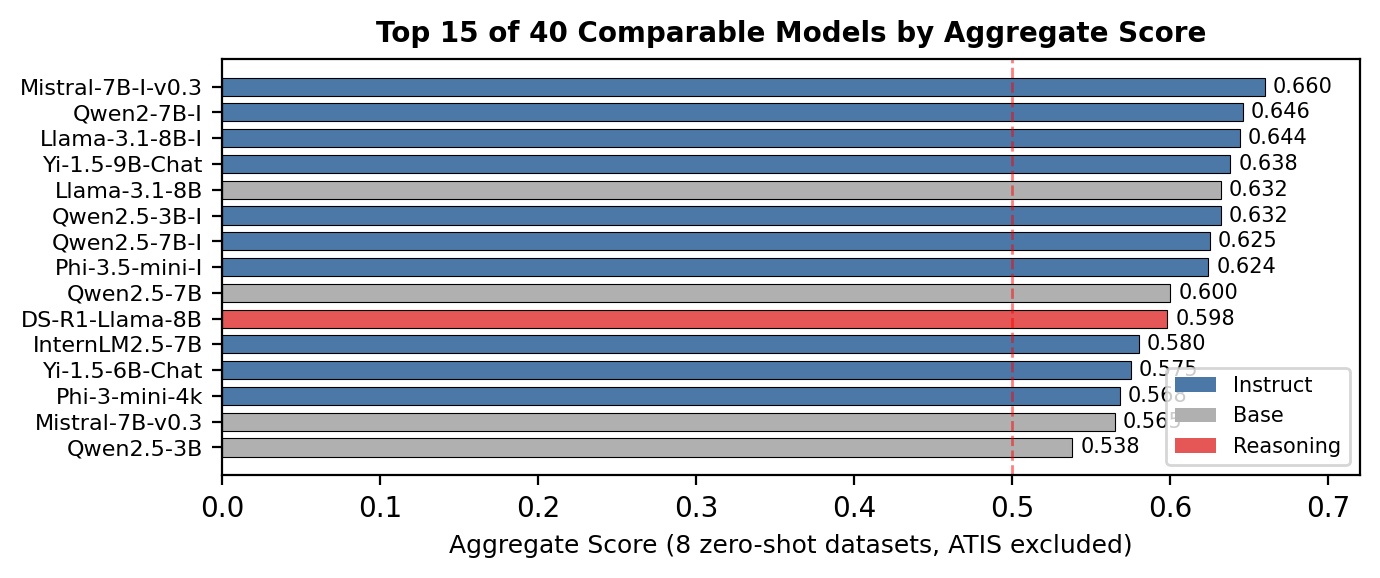}
\caption{Top 15 of the 40 comparable models ranked by group-balanced
aggregate score over eight zero-shot datasets (ATIS excluded; see
text); the complete ranking is in the Supplementary
Material.
Blue bars = Instruct models, Gray bars = Base models, Orange bars =
Reasoning models. Dashed vertical line = 0.5 reference line (an
illustrative performance regime marker, not a validated deployment
criterion).}
\label{fig:rankings}
\end{figure*}

We evaluate across eight zero-shot intent classification
datasets: CLINC150, Banking77, and SNIPS (averaged into a standard IC
composite), MASSIVE, MTOP, and the three HINT3 production
datasets (Curekart, Powerplay11, Sofmattress). A ninth dataset, ATIS,
uses five labeled demonstrations before the target query---unlike
every other dataset in this study---so ATIS reflects 5-shot in-context
performance rather than zero-shot. We therefore exclude ATIS entirely
from the aggregate score, from Figure~\ref{fig:rankings}, from all
ranking and instruction-tuning comparisons, and from the saturation
analysis; its 5-shot results are reported separately in the
Supplementary Material for reference only.
One additional model, Qwen2.5-1.5B-Instruct, is excluded from all
40-model comparisons in this section: its Banking77 evaluation used
the full test split rather than the first-500-example subset used for
every other model, making it not directly comparable (Supplementary
Material).

Figure~\ref{fig:rankings} presents the top 15 of the 40 comparable
models ranked by
their group-balanced aggregate score (Evaluation Framework section) over these eight
datasets (the complete 40-model ranking table is in the Supplementary
Material). Mistral-7B-Instruct-v0.3 leads the 40 comparable models
with an aggregate score of 0.660. Qwen2-1.5B-Instruct is the only
sub-3B model exceeding the 0.5 reference line, with a score of 0.512.
Five of the six highest-ranked models are in the 7--9B range, while
Qwen2.5-3B-Instruct has the same reported three-decimal score (0.632)
as the fifth-ranked model, Llama-3.1-8B. Removing ATIS
swaps ranks 2 and 3 relative to a naive nine-dataset aggregate
(Qwen2-7B-Instruct moves above Llama-3.1-8B-Instruct), since
Llama-3.1-8B-Instruct's ATIS score (0.846) was substantially higher
than Qwen2-7B-Instruct's (0.622); the top-ranked model is unchanged.
The presence
of a base model---Llama-3.1-8B without instruction tuning---at rank 5,
outperforming several 7B instruct models, is notable and motivates the
instruction tuning analysis in the Instruction Tuning vs.\ Parameter Scale subsection. Throughout, the 0.5
line shown in Figure~\ref{fig:rankings} (and the corresponding
parameter-scale figure in the Supplementary Material) is an
illustrative reference for the practitioner-relevant performance
regime, not a universal deployment criterion.

\finding{Finding-1:}{Mistral-7B-Instruct-v0.3 leads all 40 comparable
models at an 8-dataset zero-shot aggregate score of 0.660. Five of the
top six are 7--9B parameter models; Qwen2.5-3B-Instruct (3B) ties
the fifth-ranked model at the reported three-decimal precision (0.632).
Qwen2-1.5B-Instruct is the only sub-3B model exceeding the 0.5
reference line (0.512). A base
model (Llama-3.1-8B) at rank 5 outperforms multiple 7B instruct
models, suggesting that pretraining quality can compensate for the
absence of instruction tuning at this parameter scale.}

\subsection{Dataset-Group Performance Heatmap}

The full 40$\times$6 accuracy heatmap (Supplementary Material; a
dataset-group heatmap, not all eight zero-shot datasets shown
individually) uses the same 8-dataset zero-shot aggregate as
the full model ranking table in the Supplementary Material (ATIS excluded, Qwen2.5-1.5B-Instruct
omitted). It shows
patterns aggregate rankings alone cannot capture: the IC composite column is
predominantly green (consistent with the saturation of SNIPS
individually, see Benchmark Saturation Analysis above), Curekart
shows the widest color variation among the individually-shown
columns (Benchmark Saturation Analysis subsection), and
DeepSeek-R1-Distill-Qwen-1.5B is uniformly red (the reasoning-budget
protocol issue noted above); Powerplay11 is uniformly the lowest-accuracy
column, consistent with it being the hardest rather than most
discriminating dataset.

\subsection{Instruction Tuning vs.\ Parameter Scale}
\label{sec:instr_tuning}

\noindent\textbf{RQ-2:} \textit{Does parameter scale consistently
predict zero-shot IC performance, or does instruction tuning dominate?}

Plotting aggregate score against parameter count for the 40 comparable models
(Supplementary Material) reveals a clear pattern within the
evaluated model population: instruction tuning is a stronger
performance signal than parameter scale at the sub-9B level. Most visibly,
Qwen2.5-3B-Instruct (0.632, 3B parameters) surpasses every evaluated 7B-class base model, including Qwen2.5-7B (0.600) and Mistral-7B-v0.3 (0.565). Across the six cleanly comparable base--instruct pairs in matched families (Qwen2.5-1.5B and its instruct variant excluded as a protocol mismatch; see the Supplementary Material), instruction tuning improves the aggregate in all six (+0.012 to +0.095); we do not observe a reversal or tie in any pair. No base model below 7B
parameters outperforms any 7B+ instruct model in the top 10.

\finding{Finding-2:}{Within the matched families evaluated,
instruction tuning outweighs moderate parameter-scale increases at
sub-9B. A 3B instruct model (Qwen2.5-3B-Instruct, 0.632)
outperforms multiple 7B base models. Across all six cleanly comparable within-family pairs (one pair excluded for a protocol mismatch), instruction tuning improves the aggregate in every case (+0.012 to +0.095); no reversal or tie is observed.
Scale alone is not a reliable predictor of zero-shot IC performance.}

\subsection{Benchmark Saturation Analysis}

\noindent\textbf{RQ-3:} \textit{Have widely used IC benchmarks been
effectively solved by current open-weight models, and if so, which
should be de-emphasized?}

A benchmark is useful for model selection only if it discriminates
among the models being evaluated. We define saturation as occurring
when more than 50\% of evaluated models exceed 80\% accuracy---the
point at which the benchmark no longer meaningfully ranks models of
the current generation. Statistics for all nine evaluated datasets
are reported in the Supplementary Material; ATIS is shown for
reference but excluded from the zero-shot saturation conclusions.
Key results are
summarized below.

SNIPS is severely saturated: 32 of 41 models exceed 80\% accuracy, including models that perform very poorly elsewhere, such as BLOOM-7B (91.4\% on SNIPS, aggregate 0.208) and Qwen2.5-0.5B (85.2\%, aggregate 0.220). This level of saturation
plausibly reflects SNIPS's extremely small label space (7 intents)
and single-domain focus (smart home), which greatly reduce the
discriminative demands of the task.
ATIS is excluded from this analysis and from all aggregate scores in this paper (Overall Model Rankings subsection). In stark contrast, no model exceeds 80\% on Banking77, Sofmattress, MASSIVE, Curekart, MTOP, or Powerplay11; on Powerplay11 the best model reaches only 53.7\%, the lowest ceiling of any dataset in this suite.

\finding{Finding-3:}{SNIPS no longer discriminates among current
open-weight LLMs (32/41 models exceed 80\%) and should be
de-emphasized as a primary IC benchmark, retaining value mainly for
historical comparability. ATIS is excluded from this study's
zero-shot aggregate entirely (Overall Model Rankings subsection).
Curekart and Sofmattress show the widest top-to-bottom accuracy
ranges across models, while Powerplay11 is the hardest dataset (0/41
models exceed 80\%, best model only 53.7\%), with the lowest average
accuracy and lowest best-model accuracy.}

\subsection{Statistical Significance of Rankings}

\noindent\textbf{RQ-4:} \textit{Are the observed ranking differences
between top models statistically meaningful on a single dataset?}

We compute 95\%
confidence intervals on MASSIVE accuracy using a normal (Wald)
approximation, and apply McNemar's test for
pairwise significance. All top-5 model confidence intervals overlap
substantially (ranking by MASSIVE accuracy differs slightly from the
group-balanced aggregate ranking, Supplementary Material). All 10
pairwise comparisons among the top-5 models are non-significant
($p>0.05$), ranging from $p=0.1422$ (Qwen2-7B-Instruct vs.\
Llama-3.1-8B) to $p=0.3384$ (Qwen2.5-7B vs.\ Qwen2-7B-Instruct).
This result motivates evaluation across multiple datasets rather than
relying on a single benchmark; it does not by itself establish that
our specific group-balanced aggregate is statistically optimal (see
Limitations). The CI figure is in the Supplementary Material.

\finding{Finding-4:}{On MASSIVE, all 10 pairwise comparisons
among the top-5 models are non-significant
($p$$>$0.05, McNemar's test, $n$=500), with $p$-values ranging from
0.1422 to 0.3384, indicating that this benchmark alone cannot reliably
distinguish the leading models. This result motivates evaluation across multiple
datasets rather than relying on a single benchmark.}

\subsection{Systematic Error Patterns}

\noindent\textbf{RQ-5:} \textit{What systematic error patterns
emerge across models, and do they reflect model failure or dataset
structure?}

We analyze per-example predictions across all 41 models on MASSIVE
(500 samples), classifying each error as CROSS\_DOMAIN (the predicted
intent belongs to a different semantic domain), SAME\_DOMAIN (correct
domain but wrong intent), or EMPTY\_PRED (no recognizable intent in
the generated output) to identify the most frequent confusion pairs.
The corresponding figure
(Supplementary Material) shows the prevalence of the single most
common confusion pair identified through this analysis, not a full
per-category breakdown across all 41 models.

The most striking pattern is the \textit{play music}$\to$\textit{music query}
confusion: it is the single most common error---out of hundreds of
possible confusion pairs---for 21 of 41 models (51\%), and appears
among the top-three most frequent confusions for 26 of 41 models
(63\%), spanning diverse model families (Yi, DeepSeek-R1, Llama,
Mistral, Qwen, Phi, InternLM, ChatGLM, StableLM, TinyLlama),
parameter scales, and architectural designs. The cross-model
consistency of this pattern makes a model-specific cause unlikely: no
single architecture or training procedure is responsible. Instead, the
pattern is consistent with a structural ambiguity in MASSIVE's label
scheme---both \textit{play music} and \textit{music query} refer to
music-related actions, and zero-shot models frequently conflate them
without in-context examples to demonstrate the distinction.

\finding{Finding-5:}{\textit{Play music}$\to$\textit{music query}
confusion is the single most common error for 21/41 evaluated models
(51\%), and appears among the top-three confusions for 26/41 (63\%).
This cross-family, cross-scale consistency is consistent with
overlapping label semantics or insufficient label definitions rather
than a model-family-specific weakness. Future evaluations
of MASSIVE could provide explicit label definitions in the prompt or
report a merged-label sensitivity analysis alongside the original
benchmark evaluation.}

\paragraph{A protocol note on reasoning-distilled models.}
DeepSeek-R1-Distill-Qwen-1.5B scores 0.000 on MASSIVE; inspection of
its generated output shows this is a token-budget artifact, not a
classification-ability finding---the model's
\texttt{<think>...</think>} reasoning trace consumes the full
20-token generation budget before an answer is produced. The larger
DeepSeek-R1-Distill-Llama-8B does not show this failure (0.686), so
the issue is specific to this model/budget combination. DeepSeek-R1-Distill-Qwen-1.5B
is retained in the ranking because all models were evaluated under
the same fixed 20-token deployment protocol; its near-zero score
primarily reflects incompatibility between its reasoning-token
requirements and this output budget and should not be interpreted as
a general measure of its classification ability. See the
Supplementary Material for implementation detail.

\subsection{Deployment Efficiency}

\noindent\textbf{RQ-6:} \textit{What is the accuracy-efficiency
tradeoff for deployment, and which models offer the best
practical value?}

The full deployment efficiency table (Supplementary Material) reports
metrics for the 40 standard-protocol
models (Qwen2.5-1.5B-Instruct excluded): parameter count, peak GPU memory during inference (parsed
from vLLM's own load logs), mean inference latency (parsed from
per-model throughput logs), aggregate accuracy, and Acc/B (aggregate
score per billion parameters). We additionally compute the
\emph{Pareto-optimal} subset: the models for which no other model in
the study is simultaneously at least as accurate, at least as fast,
and at least as parameter-efficient. A model outside this set is
\emph{dominated}---some other model matches or beats it on every
axis, making it strictly harder to justify choosing.

Among the 40 models evaluated under our standard protocol
(excluding Qwen2.5-1.5B-Instruct, whose Banking77 score used a
different protocol and is not strictly comparable; see the
Supplementary Material), eleven are Pareto-optimal:
Qwen2.5-0.5B, Qwen2.5-0.5B-Instruct,
SmolLM2-135M-Instruct, SmolLM2-360M-Instruct, Qwen2-1.5B-Instruct,
Qwen2.5-1.5B, Llama-3.2-1B-Instruct,
Qwen2.5-3B, Qwen2.5-3B-Instruct,
Mistral-7B-Instruct-v0.3, and Llama-3.1-8B-Instruct. The frontier is
dominated by small models, as expected---parameter count and latency
are strongly correlated, so most Pareto-optimal points are the
smallest model that clears a given accuracy level. Only two large
models earn a place on the frontier purely on accuracy:
Mistral-7B-Instruct-v0.3 (0.660 aggregate, the highest of any
evaluated model) and Llama-3.1-8B-Instruct (0.644). Among sub-4B
models, Qwen2.5-3B-Instruct achieves the highest aggregate accuracy
(0.632) and remains Pareto-optimal, requiring 5.8GB VRAM with a
measured latency of 20.8ms/sample.

Measured directly, Qwen2-7B-Instruct does
\emph{not} offer lower latency than Mistral-7B-Instruct-v0.3 at
comparable accuracy: Qwen2-7B-Instruct is slower
(55.0ms/sample) than Mistral-7B-Instruct-v0.3 (48.5ms/sample) despite
lower accuracy (0.646 vs.\ 0.660), making Mistral-7B-Instruct-v0.3 the
better choice on both axes at 7--8B scale among instruction-tuned
models.

\finding{Finding-6:}{Eleven of the 40 models evaluated under our
standard protocol are Pareto-optimal on
accuracy/latency/parameters; the frontier is dominated by small
models, with Mistral-7B-Instruct-v0.3 and Llama-3.1-8B-Instruct the
only large models earning a place purely on accuracy. Among sub-4B
models, Qwen2.5-3B-Instruct achieves the highest aggregate accuracy
(0.632) and remains Pareto-optimal (20.8ms/sample, 5.8GB VRAM); it
does not have the highest accuracy-per-parameter efficiency in this
range (Qwen2.5-0.5B-Instruct: Acc/B=0.548).
Measured latency does not support a claim that Qwen2-7B-Instruct is
faster than Mistral-7B-Instruct-v0.3---the opposite holds
(55.0ms vs.\ 48.5ms/sample)---so Mistral-7B-Instruct-v0.3 dominates
Qwen2-7B-Instruct on both accuracy and latency among 7--8B
instruction-tuned models.}

\subsection{Robustness to Input Perturbations}

\noindent\textbf{RQ-7:} \textit{How robust are top models to realistic
input noise encountered in production environments?}

Production dialogue systems routinely receive imperfect input: users
make typos, omit punctuation, or type in all lowercase. We test all
41 models under four input perturbations on MASSIVE (500 samples,
single random realization, seed 0):
character-level typos (random substitutions, insertions, deletions),
all-lowercase conversion, punctuation removal, and random adjacent
word swap; values below are percentage-point drops in accuracy,
not relative change. We focus discussion on the 28 models with
competitive clean accuracy ($>$0.3 on MASSIVE), since near-random
models produce noisy, uninformative perturbation deltas. The
corresponding figure (Supplementary Material)
shows the full results.

Among the 28 models with competitive clean accuracy ($>$0.3 on
MASSIVE), lowercasing causes exactly 0.0 percentage points of degradation for every model
(mean and max both 0.0 percentage points) and punctuation removal causes a
negligible, statistically noisy effect (mean $-$0.24 percentage points, range $-$1.9
to +0.8 percentage points)---both are non-issues for zero-shot IC. Random adjacent
word swap has a small observed effect across the evaluated models (mean 2.2 percentage points
of degradation, median 2.3 percentage points, up to 6.4 percentage points), a perturbation not reported
in earlier analysis of this evaluation. Typos are by far the most
damaging perturbation and the only one showing wide model-to-model
variation: Qwen2.5-7B-Instruct degrades by only 1.3 percentage points
in this single-seed realization, roughly
5--20$\times$ lower than most comparable models---Llama-3.1-8B-Instruct
(6.9 points), Qwen2-7B-Instruct (11.7 points), and DeepSeek-R1-Distill-Qwen-7B
(25.9 points) among them. Critically, raw accuracy on clean input does not
predict typo robustness: Qwen2-7B-Instruct achieves the highest clean
MASSIVE accuracy (0.718) yet degrades nearly 9$\times$ more than
Qwen2.5-7B-Instruct under typos.

\finding{Finding-7:}{Among the 28 models with competitive clean
accuracy ($>$0.3 on MASSIVE), typos are the
dominant and most variable input perturbation for zero-shot IC;
lowercasing causes no measurable degradation and punctuation removal
is negligible, while random word swap has a small but consistent
effect (mean +2.2 percentage points) not previously reported. Qwen2.5-7B-Instruct
has the lowest observed typo degradation in this single-seed
realization (1.3 points) versus 6.9--26 points
for other competitive models; we did not test multiple perturbation
seeds, so this should be read as an observed result under one
realization rather than a fully general robustness guarantee. Higher raw accuracy does not predict
typo robustness---Qwen2-7B-Instruct, despite the highest clean
MASSIVE accuracy, degrades nearly 9$\times$ more than
Qwen2.5-7B-Instruct.}

\subsection{Confidence Calibration}
\label{sec:calib}

\noindent\textbf{RQ-8:} \textit{How well do model confidence scores
reflect actual prediction accuracy, and how does instruction tuning
affect calibration?}

We compute Expected Calibration Error (ECE) and a binary Brier score
for correctness confidence (the squared gap between the model's
sequence-level confidence and whether its prediction was actually
correct, not a full multiclass Brier score over the label
distribution) from free-text generation logprobabilities on MASSIVE
(500 samples), scoped to this dataset for the reasons given above.
ECE measures the
mean absolute gap between predicted confidence and actual accuracy
across probability bins; a perfectly calibrated model has ECE=0. Two
caveats apply throughout: the summed sequence-level logprobability
we use as a confidence proxy is sensitive to output label length, and
a model that is very inaccurate but consistently unconfident can
still obtain a low ECE---so ``best calibrated'' should not be read as
``best model.'' Full per-model results are in the Supplementary
Material.

Instruction tuning's effect on calibration is \emph{inconsistent}
rather than uniformly negative: across eight comparable base--instruct
pairs, five show the instruct variant less calibrated (higher ECE)
while three show it \emph{more} calibrated, and two of those reversals
are substantial (Qwen2.5-3B: base ECE=0.220 vs.\ instruct
ECE=0.085; Qwen2.5-7B: base ECE=0.173 vs.\ instruct ECE=0.066).
Llama-3.1-8B likewise reverses the pattern (base ECE=0.396 vs.\
instruct ECE=0.329). Among models with competitive accuracy
($>$0.5), Qwen2.5-7B-Instruct has the lowest ECE under our
approximate free-text confidence measure (ECE=0.066,
accuracy 0.624). SmolLM2-1.7B-Instruct shows the highest observed
ECE and substantial overconfidence under this confidence
approximation (ECE=0.768), the worst calibration in the full 41-model set.

\finding{Finding-8:}{Instruction tuning's effect on calibration is
inconsistent in this evaluation: across eight comparable
base--instruct pairs, five show worse calibration (higher ECE) for
the instruct variant and three show better calibration, with two
reversals substantial (Qwen2.5-3B and Qwen2.5-7B, where the instruct
variant roughly halves ECE relative to base). We do not find a
general calibration cost from instruction tuning. Among competitive
models, Qwen2.5-7B-Instruct has the lowest ECE under our approximate
free-text confidence measure (ECE=0.066, accuracy
0.624); SmolLM2-1.7B-Instruct is worst (ECE=0.768). Full calibration
results in the Supplementary Material.}

\section{Discussion}

\paragraph{Deployment Recommendations.}
For maximum accuracy, Mistral-7B-Instruct-v0.3 (0.660) is recommended;
for typo-prone input, Qwen2.5-7B-Instruct is preferable (1.3 vs.\
6.9--25.9 percentage points of degradation for other competitive
models, single-seed observation). Under compute
constraints, Qwen2.5-3B-Instruct (0.632, 20.8ms/sample, 5.8GB VRAM)
achieves the highest accuracy among sub-4B models while remaining
Pareto-optimal; models such as Qwen2.5-0.5B-Instruct offer higher
accuracy-per-parameter (Acc/B) if raw accuracy is less critical than
minimizing footprint. Qwen2.5-7B-Instruct has the lowest
observed ECE among competitive models under our approximate free-text
confidence measure (ECE=0.066); we did not evaluate selective
prediction, risk-coverage curves, or threshold-based abstention, so
we do not make a specific recommendation for confidence-thresholding
deployments---instruction tuning also does not reliably predict
calibration (Finding-8), so calibration should be checked directly
per model. Reasoning-distilled models require sufficient output token
budget to complete their reasoning trace before an answer can be
parsed; under a constrained budget they may fail entirely regardless
of underlying classification ability (see Implementation Details,
Supplementary Material). SNIPS should be
de-emphasized in favor of MASSIVE, MTOP, and Powerplay11.

\paragraph{Relationship to IntentGrasp.}
IntentGrasp~\cite{yin2026intentgrasp} and our work are complementary:
IntentGrasp evaluates a broad population (including proprietary
frontier systems) on unified multiple-choice intent understanding
across 49 datasets; we evaluate deployment-scale open-weight models
on single-label task-oriented IC with calibration, robustness, and
efficiency analyses absent from prior work. Scores are not directly
comparable given differing task format and metrics.

\paragraph{Limitations.}
\label{sec:limitations}
Main accuracy uses the first 500 test
examples per dataset (complete test sets for HINT3);
full-test-set evaluation may shift rankings marginally, and
index-based selection, while identical across models, is not a random
sample. Examination of the ATIS subset found a real, measurable skew: the
majority class (\texttt{flight}) is 70.8\% of the full 893-example test
set but 74.6\% of the first-500 subset (+3.8 points), so ATIS accuracy
figures in this paper are likely modestly inflated by
over-representation of the easiest class relative to the full test
set. We did not verify subset representativeness for the other five
non-HINT3 datasets. Calibration analysis is restricted to MASSIVE
based on its comparatively compact label
space; all nine datasets, including MASSIVE, use free-text generation
rather than constrained decoding, so MASSIVE's logprobabilities are an
approximate confidence proxy rather than an exact label-conditional
probability. Perturbation
robustness results use a single random realization (seed 0). The aggregate score is
group-balanced (Evaluation Framework section) rather than a simple mean across all eight
zero-shot
datasets; a sensitivity check across the 40 comparable models (Qwen2.5-1.5B-Instruct
excluded) finds the two weightings agree closely: Spearman
$\rho=0.995$ ($p<10^{-38}$), 4 of the top 5 models identical, and the
top-ranked model (Mistral-7B-Instruct-v0.3) unchanged under either
weighting. Our specific weighting choice does not appear to be
driving the paper's central ranking conclusions. All evaluations are
English-only; multilingual evaluation is deferred to future work. We
evaluate single-label classification only; multi-label intent
detection is outside our scope.

\section{Conclusion}

We presented a systematic evaluation of 41 open-weight
LLMs across eight zero-shot intent classification datasets and one
five-shot ATIS evaluation, targeting the
sub-9B deployment space that prior frontier-model evaluations overlook.
Our results show that instruction tuning can outweigh moderate
increases in parameter scale within matched model families, that
rankings based on one benchmark may not significantly distinguish
leading models, and that robustness, calibration, and efficiency
expose practical tradeoffs not captured by aggregate accuracy alone.
Taken together, these
findings---detailed in the Results and Analysis section---provide actionable,
evidence-based guidance for practitioners selecting open-weight models
for production intent classification systems.

\bibliography{references}

\begin{thebibliography}{35}
\providecommand{\natexlab}[1]{#1}

\bibitem[{Abdin et~al.(2024)Abdin, Jacobs, Awan, Aneja, Awadallah, Awadalla,
  Bach, Bahree, Bakhtiari, Behl et~al.}]{phi3_2024}
Abdin, M.; Jacobs, S.~A.; Awan, A.~A.; Aneja, J.; Awadallah, A.; Awadalla, H.;
  Bach, N.; Bahree, A.; Bakhtiari, A.; Behl, H.; et~al. 2024.
\newblock Phi-3 Technical Report: A Highly Capable Language Model Locally on
  Your Phone.
\newblock \emph{arXiv preprint arXiv:2404.14219}.

\bibitem[{Allal et~al.(2025)Allal, Lozhkov, Bakouch, Blazquez, Penedo,
  Tunstall, Marafioti, Kydlicek, Lajarin, Srivastav et~al.}]{smollm2_2025}
Allal, L.~B.; Lozhkov, A.; Bakouch, E.; Blazquez, G.~M.; Penedo, G.; Tunstall,
  L.; Marafioti, A.; Kydlicek, H.; Lajarin, A.~P.; Srivastav, V.; et~al. 2025.
\newblock SmolLM2: When Smol Goes Big -- Data-Centric Training of a Small
  Language Model.
\newblock \emph{arXiv preprint arXiv:2502.02737}.

\bibitem[{Bellagente et~al.(2024)Bellagente, Tow, Mahan, Phung, Zhuravinskyi,
  Adithyan, Baicoianu, Brooks, Cooper, Datta et~al.}]{stablelm2_2024}
Bellagente, M.; Tow, J.; Mahan, D.; Phung, D.; Zhuravinskyi, M.; Adithyan, R.;
  Baicoianu, J.; Brooks, B.; Cooper, N.; Datta, A.; et~al. 2024.
\newblock Stable LM 2 1.6B Technical Report.
\newblock \emph{arXiv preprint arXiv:2402.17834}.

\bibitem[{Biderman et~al.(2023)Biderman, Schoelkopf, Anthony, Bradley, O'Brien,
  Hallahan, Khan, Purohit, Prashanth, Raff, Skowron, Sutawika, and van~der
  Wal}]{pythia_2023}
Biderman, S.; Schoelkopf, H.; Anthony, Q.~G.; Bradley, H.; O'Brien, K.;
  Hallahan, E.; Khan, M.~A.; Purohit, S.; Prashanth, U.~S.; Raff, E.; Skowron,
  A.; Sutawika, L.; and van~der Wal, O. 2023.
\newblock Pythia: A Suite for Analyzing Large Language Models Across Training
  and Scaling.
\newblock In \emph{International Conference on Machine Learning}, 2397--2430.
  PMLR.

\bibitem[{{BigScience Workshop} et~al.(2022){BigScience Workshop}, Le~Scao,
  Fan, Akiki, Pavlick, Ili{\'c}, Hesslow, Castagn{\'e}, Luccioni, Yvon
  et~al.}]{bloom_2022}
{BigScience Workshop}; Le~Scao, T.; Fan, A.; Akiki, C.; Pavlick, E.; Ili{\'c},
  S.; Hesslow, D.; Castagn{\'e}, R.; Luccioni, A.~S.; Yvon, F.; et~al. 2022.
\newblock BLOOM: A 176B-Parameter Open-Access Multilingual Language Model.
\newblock \emph{arXiv preprint arXiv:2211.05100}.

\bibitem[{Brown et~al.(2020)Brown, Mann, Ryder, Subbiah, Kaplan, Dhariwal,
  Neelakantan, Shyam, Sastry, Askell et~al.}]{brown2020gpt3}
Brown, T.; Mann, B.; Ryder, N.; Subbiah, M.; Kaplan, J.~D.; Dhariwal, P.;
  Neelakantan, A.; Shyam, P.; Sastry, G.; Askell, A.; et~al. 2020.
\newblock Language models are few-shot learners.
\newblock In \emph{Advances in Neural Information Processing Systems},
  volume~33, 1877--1901.

\bibitem[{Cai et~al.(2024)Cai, Cao, Chen, Chen, Chen, Chen, Chen, Chen, Chen,
  Chu et~al.}]{internlm2_2024}
Cai, Z.; Cao, M.; Chen, H.; Chen, K.; Chen, K.; Chen, X.; Chen, X.; Chen, Z.;
  Chen, Z.; Chu, P.; et~al. 2024.
\newblock InternLM2 Technical Report.
\newblock \emph{arXiv preprint arXiv:2403.17297}.

\bibitem[{Casanueva et~al.(2020)Casanueva, Te{\v{m}}{\v{c}}inas, Gerz,
  Henderson, and Vuli{\'c}}]{casanueva2020banking77}
Casanueva, I.; Te{\v{m}}{\v{c}}inas, T.; Gerz, D.; Henderson, M.; and
  Vuli{\'c}, I. 2020.
\newblock Efficient intent detection with dual sentence encoders.
\newblock In \emph{Proceedings of the 2nd Workshop on NLP for Conversational
  AI}.

\bibitem[{Coucke et~al.(2018)Coucke, Saade, Ball, Bluche, Caulier, Leroy,
  Doumouro, Gisselbrecht, Caltagirone, Lavril et~al.}]{coucke2018snips}
Coucke, A.; Saade, A.; Ball, A.; Bluche, T.; Caulier, A.; Leroy, D.; Doumouro,
  C.; Gisselbrecht, T.; Caltagirone, F.; Lavril, T.; et~al. 2018.
\newblock Snips voice platform: an embedded spoken language understanding
  system for private-by-design voice interfaces.
\newblock \emph{arXiv preprint arXiv:1805.10190}.

\bibitem[{{DeepSeek-AI} et~al.(2025){DeepSeek-AI}, Guo, Yang, Zhang, Song,
  Zhang, Xu, Zhu, Ma, Wang et~al.}]{deepseekr1_2025}
{DeepSeek-AI}; Guo, D.; Yang, D.; Zhang, H.; Song, J.; Zhang, R.; Xu, R.; Zhu,
  Q.; Ma, S.; Wang, P.; et~al. 2025.
\newblock DeepSeek-R1: Incentivizing Reasoning Capability in LLMs via
  Reinforcement Learning.
\newblock \emph{arXiv preprint arXiv:2501.12948}.

\bibitem[{FitzGerald et~al.(2022)FitzGerald, Hench, Peris, Mackie, Rottmann,
  Sanchez, Nash, Urbach, Kakarala, Singh et~al.}]{fitzgerald2022massive}
FitzGerald, J.; Hench, C.; Peris, C.; Mackie, S.; Rottmann, K.; Sanchez, A.;
  Nash, A.; Urbach, L.; Kakarala, V.; Singh, R.; et~al. 2022.
\newblock MASSIVE: A 1M-example multilingual natural language understanding
  dataset with 51 typologically-diverse languages.
\newblock \emph{arXiv preprint arXiv:2204.08582}.

\bibitem[{Grattafiori et~al.(2024)Grattafiori, Dubey, Jauhri, Pandey, Kadian,
  Al-Dahle, Letman, Mathur, Schelten, Vaughan et~al.}]{llama3_2024}
Grattafiori, A.; Dubey, A.; Jauhri, A.; Pandey, A.; Kadian, A.; Al-Dahle, A.;
  Letman, A.; Mathur, A.; Schelten, A.; Vaughan, A.; et~al. 2024.
\newblock The Llama 3 Herd of Models.
\newblock \emph{arXiv preprint arXiv:2407.21783}.

\bibitem[{Guo et~al.(2017)Guo, Pleiss, Sun, and
  Weinberger}]{guo2017calibration}
Guo, C.; Pleiss, G.; Sun, Y.; and Weinberger, K.~Q. 2017.
\newblock On calibration of modern neural networks.
\newblock In \emph{Proceedings of the 34th International Conference on Machine
  Learning}.

\bibitem[{Hemphill, Godfrey, and Doddington(1990)}]{hemphill1990atis}
Hemphill, C.~T.; Godfrey, J.~J.; and Doddington, G.~R. 1990.
\newblock The {ATIS} spoken language systems pilot corpus.
\newblock In \emph{Speech and Natural Language: Proceedings of a Workshop}.

\bibitem[{Hu et~al.(2024)Hu, Tu, Han, He, Cui, Long, Zheng, Fang, Huang, Zhao
  et~al.}]{minicpm_2024}
Hu, S.; Tu, Y.; Han, X.; He, C.; Cui, G.; Long, X.; Zheng, Z.; Fang, Y.; Huang,
  Y.; Zhao, W.; et~al. 2024.
\newblock MiniCPM: Unveiling the Potential of Small Language Models with
  Scalable Training Strategies.
\newblock \emph{arXiv preprint arXiv:2404.06395}.

\bibitem[{Jiang et~al.(2023)Jiang, Sablayrolles, Mensch, Bamford, Chaplot,
  de~las Casas, Bressand, Lengyel, Lample, Saulnier et~al.}]{mistral7b_2023}
Jiang, A.~Q.; Sablayrolles, A.; Mensch, A.; Bamford, C.; Chaplot, D.~S.; de~las
  Casas, D.; Bressand, F.; Lengyel, G.; Lample, G.; Saulnier, L.; et~al. 2023.
\newblock Mistral 7B.
\newblock \emph{arXiv preprint arXiv:2310.06825}.

\bibitem[{Kwon et~al.(2023)Kwon, Li, Zhuang, Sheng, Zheng, Yu, Gonzalez, Zhang,
  and Stoica}]{kwon2023vllm}
Kwon, W.; Li, Z.; Zhuang, S.; Sheng, Y.; Zheng, L.; Yu, C.~H.; Gonzalez, J.~E.;
  Zhang, H.; and Stoica, I. 2023.
\newblock Efficient memory management for large language model serving with
  pagedattention.
\newblock In \emph{Proceedings of the 29th Symposium on Operating Systems
  Principles}.

\bibitem[{Larson and Leach(2022)}]{larson2022survey}
Larson, S.; and Leach, K. 2022.
\newblock A survey of intent classification and slot-filling datasets for
  task-oriented dialog.
\newblock \emph{arXiv preprint arXiv:2207.13211}.

\bibitem[{Larson et~al.(2019)Larson, Mahendran, Peper, Clarke, Lee, Hill,
  Kummerfeld, Leach, Laurenzano, Tang et~al.}]{larson2019clinc}
Larson, S.; Mahendran, A.; Peper, J.~J.; Clarke, C.; Lee, A.; Hill, P.;
  Kummerfeld, J.~K.; Leach, K.; Laurenzano, M.~A.; Tang, L.; et~al. 2019.
\newblock An evaluation dataset for intent classification and out-of-scope
  prediction.
\newblock In \emph{Proceedings of EMNLP-IJCNLP}.

\bibitem[{Li et~al.(2025)Li, Lyu, Yang et~al.}]{li2025consint}
Li, X.; Lyu, T.; Yang, S.; et~al. 2025.
\newblock Consintbench: Evaluating language models on real-world consumer
  intent understanding.
\newblock \emph{arXiv preprint arXiv:2510.13499}.

\bibitem[{Mehri and Eric(2021)}]{mehri2021zeroshot}
Mehri, S.; and Eric, M. 2021.
\newblock Example-driven intent prediction with observers.
\newblock In \emph{Proceedings of NAACL}.

\bibitem[{Mehta et~al.(2020)Mehta, Jain, Chaturvedi, and Modi}]{mehta2020hint3}
Mehta, G.; Jain, C.; Chaturvedi, M.; and Modi, K. 2020.
\newblock {HINT3}: Raising the bar for intent detection in the wild.
\newblock In \emph{Proceedings of the First Workshop on Insights from Negative
  Results in NLP}.

\bibitem[{Park et~al.(2024)Park, Baek, Kim, Shin, and Lee}]{park2024dynamic}
Park, G.; Baek, I.; Kim, B.; Shin, J.; and Lee, H. 2024.
\newblock Dynamic Label Name Refinement for Few-Shot Dialogue Intent
  Classification.
\newblock \emph{arXiv preprint arXiv:2412.15603}.

\bibitem[{Qin et~al.(2020)Qin, Xu, Che, and Liu}]{qin2020agif}
Qin, L.; Xu, X.; Che, W.; and Liu, T. 2020.
\newblock {AGIF}: An adaptive graph-interactive framework for joint multiple
  intent detection and slot filling.
\newblock In \emph{Findings of EMNLP}.

\bibitem[{Schick and Sch{\"u}tze(2021)}]{schick2021exploiting}
Schick, T.; and Sch{\"u}tze, H. 2021.
\newblock Exploiting cloze-questions for few-shot text classification and
  natural language inference.
\newblock In \emph{Proceedings of EACL}.

\bibitem[{{Team GLM} et~al.(2024){Team GLM}, Zeng, Xu, Wang, Zhang, Yin, Zhang,
  Rojas, Feng, Zhao et~al.}]{chatglm_2024}
{Team GLM}; Zeng, A.; Xu, B.; Wang, B.; Zhang, C.; Yin, D.; Zhang, D.; Rojas,
  D.; Feng, G.; Zhao, H.; et~al. 2024.
\newblock ChatGLM: A Family of Large Language Models from GLM-130B to GLM-4 All
  Tools.
\newblock \emph{arXiv preprint arXiv:2406.12793}.

\bibitem[{Wang and Komatsuzaki(2021)}]{gptj_2021}
Wang, B.; and Komatsuzaki, A. 2021.
\newblock {GPT-J-6B: A 6 Billion Parameter Autoregressive Language Model}.
\newblock https://github.com/kingoflolz/mesh-transformer-jax.

\bibitem[{Wei et~al.(2022)Wei, Bosma, Zhao, Guu, Yu, Lester, Du, Dai, and
  Le}]{wei2022flan}
Wei, J.; Bosma, M.; Zhao, V.~Y.; Guu, K.; Yu, A.~W.; Lester, B.; Du, N.; Dai,
  A.~M.; and Le, Q.~V. 2022.
\newblock Finetuned language models are zero-shot learners.
\newblock In \emph{Proceedings of ICLR}.

\bibitem[{Yang et~al.(2024{\natexlab{a}})Yang, Yang, Hui, Zheng, Yu, Zhou, Li,
  Li, Liu, Huang et~al.}]{qwen2_2024}
Yang, A.; Yang, B.; Hui, B.; Zheng, B.; Yu, B.; Zhou, C.; Li, C.; Li, C.; Liu,
  D.; Huang, F.; et~al. 2024{\natexlab{a}}.
\newblock Qwen2 Technical Report.
\newblock \emph{arXiv preprint arXiv:2407.10671}.

\bibitem[{Yang et~al.(2024{\natexlab{b}})Yang, Yang, Zhang, Hui, Zheng, Yu, Li,
  Liu, Huang et~al.}]{qwen25_2024}
Yang, A.; Yang, B.; Zhang, B.; Hui, B.; Zheng, B.; Yu, B.; Li, C.; Liu, D.;
  Huang, F.; et~al. 2024{\natexlab{b}}.
\newblock Qwen2.5 Technical Report.
\newblock \emph{arXiv preprint arXiv:2412.15115}.

\bibitem[{Yang et~al.(2025)Yang, Wang, Xu et~al.}]{yang2025sessionintent}
Yang, Y.; Wang, W.; Xu, B.; et~al. 2025.
\newblock SessionIntentBench: A multi-task inter-session intention-shift
  modeling benchmark.
\newblock \emph{arXiv preprint arXiv:2507.20185}.

\bibitem[{Yin, Li, and Carenini(2026)}]{yin2026intentgrasp}
Yin, Y.; Li, C.; and Carenini, G. 2026.
\newblock Intentgrasp: A comprehensive benchmark for intent understanding.
\newblock \emph{arXiv preprint arXiv:2605.06832}.

\bibitem[{Young et~al.(2024)Young, Chen, Li, Huang, Zhang, Zhang, Li, Zhu,
  Chen, Chang et~al.}]{yi_2024}
Young, A.; Chen, B.; Li, C.; Huang, C.; Zhang, G.; Zhang, G.; Li, H.; Zhu, J.;
  Chen, J.; Chang, J.; et~al. 2024.
\newblock Yi: Open Foundation Models by 01.AI.
\newblock \emph{arXiv preprint arXiv:2403.04652}.

\bibitem[{Zhang et~al.(2024)Zhang, Zeng, Wang, and Lu}]{tinyllama_2024}
Zhang, P.; Zeng, G.; Wang, T.; and Lu, W. 2024.
\newblock TinyLlama: An Open-Source Small Language Model.
\newblock \emph{arXiv preprint arXiv:2401.02385}.

\bibitem[{Zhao et~al.(2021)Zhao, Wallace, Feng, Klein, and
  Singh}]{zhao2021calibrate}
Zhao, T.~Z.; Wallace, E.; Feng, S.; Klein, D.; and Singh, S. 2021.
\newblock Calibrate before use: Improving few-shot performance of language
  models.
\newblock \emph{arXiv preprint}.

\end{thebibliography}

\clearpage
\section*{Supplementary Material}
This supplementary material provides complete model rankings, extended
experimental results, prompt templates, calibration results, robustness
analyses, statistical tests, and deployment-efficiency details
supporting the main paper.

\section{Extended Related Work}
\label{app:related}

This section expands on the condensed Related Work in the main paper
with additional citations and closer comparisons to prior benchmarks.
Citations in this document use author-year form; full references are
listed in the main paper's bibliography.

\paragraph{Intent Classification Benchmarks.}
IC benchmarks span three decades: ATIS (Hemphill, Godfrey, and Doddington 1990),
SNIPS (Coucke et al. 2018) (7 intents), CLINC150 (Larson et al. 2019)
(150 intents, out-of-scope detection), Banking77 (Casanueva et al. 2020)
(77 fine-grained banking intents), MASSIVE (FitzGerald et al. 2022)
(60 intents, 18 domains, real Alexa interactions), and
MTOP (Li et al. 2021) (117 intents, hierarchical). HINT3 (Mehta et al. 2020)
introduced three real Indian e-commerce production
datasets---Curekart, Powerplay11, Sofmattress---to expose the gap
between academic benchmarks and deployment reality. Larson and
Leach (2022) surveyed IC datasets and noted growing
saturation concerns, which our analysis confirms and extends.

\paragraph{Zero-Shot IC with LLMs.}
Zero-shot classification has improved with scale (Brown et al. 2020);
Wei et al. (2022) showed instruction-tuned models
generalize well zero-shot, and Schick and
Sch\"{u}tze (2021) showed task descriptions in
prompts approach few-shot performance. For IC specifically, Mehri and
Eric (2021) showed pre-trained LMs can perform
zero-shot intent detection with appropriate prompting. Closest to our
evaluation setting, Park et al. (2024) evaluate
Qwen2.5-1.5B/7B-Instruct and Llama3-8B-Instruct on the same HINT3
datasets we use (Curekart, Powerplay11, Sofmattress) plus
Banking77/CLINC150, but in a 10-shot in-context-learning regime with
LLM-driven dynamic label refinement rather than true zero-shot; their
finding that Powerplay11 shows the largest relative gains from label
clarification (up to +5.23 points) is consistent with our own
observation (main paper's Benchmark Saturation Analysis) that Powerplay11 is the
hardest dataset in our zero-shot suite, suggesting the
same underlying label-space property (fine-grained, semantically
overlapping intents) drives both results.

\paragraph{Benchmarking LLMs for Intent Understanding.}
IntentGrasp (Yin, Li, and Carenini 2026), the most comprehensive recent
evaluation, reformats 49 intent-related datasets as multiple-choice QA
across 20 frontier LLMs (all below 60\% F1) and proposes Intentional
Fine-Tuning as a remedy. ConsintBench (Li et al. 2025) and
SessionIntentBench (Yang et al. 2025) address e-commerce and
session-level intent, but their underlying datasets are not publicly
available (see main paper's References for full citations). Our work occupies
a distinct niche: open-weight sub-9B models with deployment-oriented
analyses (calibration, robustness, efficiency, ranking reliability)
absent from all of the above.

\paragraph{Calibration of LLMs.}
Modern neural networks tend toward overconfidence (Guo et al. 2017);
Zhao et al. (2021) identified calibration failures in
GPT-3 on classification tasks. Instruction tuning's effect on
calibration has received limited attention for open-weight LLMs on IC,
which we study directly.

\section{Full Model Rankings}
\label{app:models}

Table~\ref{tab:full_models} presents 40 of the 41 evaluated models
ranked by aggregate score. \textbf{ATIS is excluded from this
aggregate}: its prompt template embeds five
labeled examples, so ATIS results are 5-shot, not zero-shot (see
Prompt Templates below), and including it would misrepresent this as
a nine-dataset zero-shot study. The aggregate score is
therefore the mean across six dataset groups computed over
\textbf{eight} zero-shot datasets: the standard IC composite (mean of
CLINC150, Banking77, SNIPS only), MASSIVE, MTOP, Curekart,
Powerplay11, and Sofmattress. ATIS's 5-shot accuracy is reported
separately in the rightmost column for reference and is not part of
any ranking, aggregate, or comparison in this paper.
Qwen2.5-1.5B-Instruct is omitted from this table entirely and
presented separately below: its Banking77 evaluation used the full
test split rather than the first-500 subset used for all other 40
models, so it is not comparable under this protocol.

\begin{table}[h]
\centering
\caption{40 of 41 evaluated models (Qwen2.5-1.5B-Instruct excluded,
see below), ranked by an 8-dataset zero-shot aggregate score. T = Type
(I=Instruct, B=Base, R=Reasoning). IC = mean accuracy across
CLINC150, Banking77, SNIPS (3 datasets; ATIS excluded from this
aggregate). ATIS col.\ = 5-shot accuracy, reported for reference only,
not part of the aggregate. Blue shading = top-5 models.}
\label{tab:full_models}
\small\setlength{\tabcolsep}{3pt}
\begin{tabular}{rlrrrrrr}
\toprule
\# & Model & T & Agg(8ds) & IC(3ds) & MASS & MTOP & ATIS(5-shot) \\
\midrule
\rowcolor{lightblue}
1 & Mistral-7B-I-v0.3 & I & .660 & .823 & .622 & .652 & (.846) \\
\rowcolor{lightblue}
2 & Qwen2-7B-I & I & .646 & .790 & .718 & .584 & (.622) \\
\rowcolor{lightblue}
3 & Llama-3.1-8B-I & I & .644 & .822 & .642 & .626 & (.846) \\
\rowcolor{lightblue}
4 & Yi-1.5-9B-Chat & I & .638 & .660 & .640 & .642 & (.678) \\
\rowcolor{lightblue}
5 & Llama-3.1-8B & B & .632 & .801 & .688 & .568 & (.888) \\
6 & Qwen2.5-3B-I & I & .632 & .763 & .650 & .588 & (.720) \\
7 & Qwen2.5-7B-I & I & .625 & .774 & .624 & .600 & (.462) \\
8 & Phi-3.5-mini-I & I & .624 & .746 & .580 & .574 & (.626) \\
9 & Qwen2.5-7B & B & .600 & .745 & .694 & .600 & (.798) \\
10 & DS-R1-Llama-8B & R & .598 & .653 & .686 & .582 & (.928) \\
11 & InternLM2.5-7B & I & .580 & .795 & .692 & .628 & (.822) \\
12 & Yi-1.5-6B-Chat & I & .575 & .763 & .590 & .578 & (.856) \\
13 & Phi-3-mini-4k & I & .568 & .621 & .594 & .460 & (.562) \\
14 & Mistral-7B-v0.3 & B & .565 & .755 & .650 & .514 & (.928) \\
15 & Qwen2.5-3B & B & .538 & .681 & .622 & .544 & (.770) \\
16 & Mistral-7B-I-v0.1 & I & .523 & .707 & .584 & .510 & (.644) \\
17 & Qwen2-1.5B-I & I & .512 & .589 & .550 & .488 & (.242) \\
18 & Llama-3.2-3B-I & I & .508 & .685 & .438 & .506 & (.314) \\
19 & Llama-3.2-3B & B & .485 & .695 & .604 & .242 & (.882) \\
20 & DS-R1-Qwen-7B & R & .463 & .517 & .518 & .452 & (.478) \\
21 & chatglm3-6b & I & .438 & .659 & .598 & .390 & (.724) \\
22 & MiniCPM-2B & I & .425 & .660 & .462 & .414 & (.772) \\
23 & Qwen2.5-1.5B & B & .416 & .577 & .496 & .428 & (.890) \\
24 & phi-2 & B & .399 & .599 & .422 & .320 & (.524) \\
25 & StableLM-1.6B & I & .359 & .499 & .462 & .280 & (.750) \\
26 & TinyLlama-1.1B & I & .279 & .573 & .358 & .072 & (.526) \\
27 & Llama-3.2-1B-I & I & .279 & .491 & .296 & .252 & (.804) \\
28 & Qwen2.5-0.5B-I & I & .274 & .479 & .342 & .292 & (.842) \\
29 & Llama-3.2-1B & B & .220 & .555 & .158 & .020 & (.760) \\
30 & Qwen2.5-0.5B & B & .205 & .461 & .234 & .178 & (.824) \\
31 & BLOOM-7B & B & .192 & .447 & .250 & .144 & (.820) \\
32 & Pythia-2.8B & B & .139 & .359 & .098 & .014 & (.760) \\
33 & Pythia-6.9B & B & .130 & .399 & .100 & .024 & (.838) \\
34 & GPT-J-6B & B & .125 & .317 & .128 & .030 & (.750) \\
35 & Phi-1.5 & B & .121 & .290 & .144 & .038 & (.758) \\
36 & SmolLM2-360M-I & I & .048 & .139 & .088 & .018 & (.832) \\
37 & Pythia-1.4B & B & .047 & .113 & .030 & .000 & (.762) \\
38 & SmolLM2-135M-I & I & .043 & .191 & .006 & .012 & (.806) \\
39 & SmolLM2-1.7B-I & I & .036 & .008 & .064 & .002 & (.546) \\
40 & DS-R1-Qwen-1.5B & R & .002 & .001 & .000 & .012 & (.858) \\
\bottomrule
\end{tabular}
\end{table}

\paragraph{Excluded model.} Qwen2.5-1.5B-Instruct achieves an
8-dataset aggregate of 0.462 (IC(3ds)=0.552, MASSIVE=0.469,
MTOP=0.529, ATIS 5-shot=0.404) under the same computation, but its
Banking77 score (0.4146) was measured on the full test split rather
than the first-500 subset used for all other models, so it is not
included in Table~\ref{tab:full_models}'s ranking or in any
cross-model comparison in this paper. It is reported here for
completeness only.

\paragraph{DeepSeek-R1-Distill-Qwen-1.5B.} This model is
retained with its numbered rank in Table~\ref{tab:full_models} because
it was evaluated under the identical protocol as all other models. Its near-zero score
is a token-budget artifact, not a classification-ability result: its
reasoning trace consumes the full 20-token generation budget before
an answer is produced (see the main paper's protocol note). It should
not be interpreted as a general measure of its classification
ability or used to draw conclusions about reasoning-model IC
performance more broadly.

\section{Calibration --- Full Results and Scope Justification}
\label{app:calib}

\paragraph{Scope Justification.} Calibration analysis requires
generation logprobabilities that represent genuine label-level
confidence. All nine datasets, including MASSIVE, use free-text
generation in this evaluation; we do not apply constrained decoding.
Generated tokens are therefore not guaranteed to correspond to a valid
label on any dataset. MASSIVE's label space is small and its labels are
short (predominantly one-to-two words), which in practice makes greedy
free-text generations align closely with the label vocabulary, so the
sequence-level logprobability of the generated text returned by the
evaluation pipeline is a usable
approximate confidence signal. The other datasets in this evaluation
(CLINC150, Banking77, ATIS, SNIPS, MTOP, HINT3) have longer or more varied label
surface forms, where free-text generations diverge from the label
vocabulary more often, making output logprobabilities a substantially
noisier calibration signal. Restricting calibration to MASSIVE is
therefore a pragmatic scope decision, not a claim of exact
label-conditional probability; see the main paper's Limitations
paragraph for the caveat this entails.

\paragraph{Implementation Details.} Confidence is obtained by
exponentiating the sequence-level log-probability returned by the
evaluation pipeline. ECE is computed with $n=10$ equal-width bins
(\texttt{np.linspace(0, 1, 11)}), confirmed directly from
\texttt{analysis/calibration.py}. Invalid generations were not
analyzed separately.

\begin{table}[h]
\centering
\caption{Full calibration results on MASSIVE (500 samples, free-text
generation; see the main paper's calibration scope discussion). ECE$\downarrow$ =
Expected Calibration Error (lower is better calibrated). T = Type
(B=Base, I=Instruct, R=Reasoning).}
\label{tab:calib_full}
\small\setlength{\tabcolsep}{3.5pt}
\begin{tabular}{lcrrr}
\toprule
\textbf{Model} & \textbf{T} & \textbf{Acc} & \textbf{ECE$\downarrow$} & \textbf{Brier$\downarrow$} \\
\midrule
Pythia-1.4B        & B & .030 & \textbf{.032} & .035 \\
Pythia-2.8B        & B & .098 & .035 & .082 \\
Pythia-6.9B        & B & .100 & .046 & .093 \\
gpt-j-6b           & B & .128 & .053 & .109 \\
Phi-1.5            & B & .144 & .072 & .130 \\
Qwen2.5-0.5B       & B & .234 & .074 & .148 \\
Qwen2.5-1.5B       & B & .496 & .098 & .189 \\
phi-2              & B & .422 & .142 & .224 \\
BLOOM-7B           & B & .250 & .152 & .210 \\
Llama-3.2-1B       & B & .158 & .157 & .153 \\
Mistral-7B-v0.3    & B & .650 & .171 & .188 \\
Qwen2.5-7B         & B & .694 & .173 & .153 \\
Qwen2.5-3B         & B & .622 & .220 & .228 \\
Llama-3.2-3B       & B & .604 & .380 & .324 \\
Llama-3.1-8B       & B & .688 & .396 & .316 \\
\midrule
SmolLM2-135M-I     & I & .006 & .041 & .011 \\
Qwen2.5-7B-I       & I & .624 & .066 & .160 \\
SmolLM2-360M-I     & I & .088 & .077 & .093 \\
internlm2.5-7b     & I & .692 & .080 & .156 \\
Qwen2.5-3B-I       & I & .650 & .085 & .212 \\
Mistral-7B-I-v0.1  & I & .584 & .091 & .163 \\
TinyLlama-1.1B     & I & .358 & .095 & .171 \\
Phi-3-mini-4k-I    & I & .594 & .110 & .162 \\
MiniCPM-2B         & I & .462 & .112 & .217 \\
Qwen2.5-0.5B-I     & I & .342 & .124 & .155 \\
Yi-1.5-6B-Chat     & I & .590 & .172 & .217 \\
Llama-3.2-1B-I     & I & .296 & .179 & .218 \\
chatglm3-6b        & I & .598 & .200 & .225 \\
Qwen2.5-1.5B-I     & I & .580 & .207 & .259 \\
stablelm-1.6b      & I & .462 & .224 & .255 \\
Yi-1.5-9B-Chat     & I & .634 & .245 & .239 \\
Mistral-7B-I-v0.3  & I & .622 & .249 & .251 \\
Phi-3.5-mini-I     & I & .580 & .262 & .263 \\
Qwen2-1.5B-I       & I & .550 & .265 & .290 \\
Llama-3.1-8B-I     & I & .642 & .329 & .305 \\
Qwen2-7B-I         & I & .718 & .335 & .278 \\
Llama-3.2-3B-I     & I & .438 & .448 & .370 \\
\rowcolor{lightred}
SmolLM2-1.7B-I     & I & .064 & .768 & .712 \\
\midrule
DS-R1-Qwen-1.5B    & R & .000 & .119 & .015 \\
DS-R1-Qwen-7B      & R & .518 & .159 & .214 \\
DS-R1-Llama-8B     & R & .686 & .191 & .200 \\
\bottomrule
\end{tabular}
\end{table}

\section{Additional Figures}
\label{app:figures}

\subsection{Benchmark Saturation Analysis (Full Table)}
\label{app:saturation}

The full saturation analysis referenced in the main paper's
Benchmark Saturation Analysis subsection is presented below.

\begin{table*}[t]
\centering
\caption{Benchmark saturation analysis. \#$>$80\% = number of models
exceeding 80\% accuracy (out of 41). Top Acc $-$ Bot Acc = the
accuracy range across models on that dataset (larger range =
more discriminating). Recommendations based
on saturation thresholds and this range.}
\label{tab:saturation}
\setlength{\tabcolsep}{6pt}
\begin{tabular}{llrrrrl}
\toprule
\textbf{Dataset} & \textbf{Domain} & \textbf{Avg Acc} & \textbf{Top Acc} & \textbf{Bot Acc} & \textbf{\#$>$80\%} & \textbf{Recommendation} \\
\midrule
\rowcolor{lightred}
SNIPS & Smart home & 0.807 & 0.998 & 0.002 & 32/41 & \textbf{De-emphasize --- saturated} \\
\rowcolor{lightyellow}
ATIS & Airline travel & 0.718 & 0.928 & 0.242 & 17/41 & Excluded (5-shot, not zero-shot) \\
CLINC150 & General assistant & 0.468 & 0.836 & 0.000 & 3/41 & Retain \\
MASSIVE & Voice assistant & 0.432 & 0.718 & 0.000 & 0/41 & Retain --- highly discriminating \\
Banking77 & Retail banking & 0.388 & 0.708 & 0.000 & 0/41 & Retain --- highly discriminating \\
Sofmattress & E-commerce & 0.378 & 0.731 & 0.000 & 0/41 & Retain --- highly discriminating \\
Curekart & E-commerce & 0.364 & 0.760 & 0.000 & 0/41 & Retain --- highly discriminating \\
MTOP & Facebook Asst. & 0.348 & 0.652 & 0.000 & 0/41 & Retain --- highly discriminating \\
Powerplay11 & Gaming & 0.299 & 0.537 & 0.000 & 0/41 & Retain --- hardest dataset \\
\bottomrule
\end{tabular}
\end{table*}

\subsection{Parameter Scale vs.\ Instruction Tuning (Figure)}
\label{app:scale_fig}

The full parameter-scale-vs-performance figure referenced in the
main paper's Instruction Tuning vs.\ Parameter Scale subsection is presented below.

\begin{figure*}[t]
\centering
\includegraphics[width=0.6\textwidth]{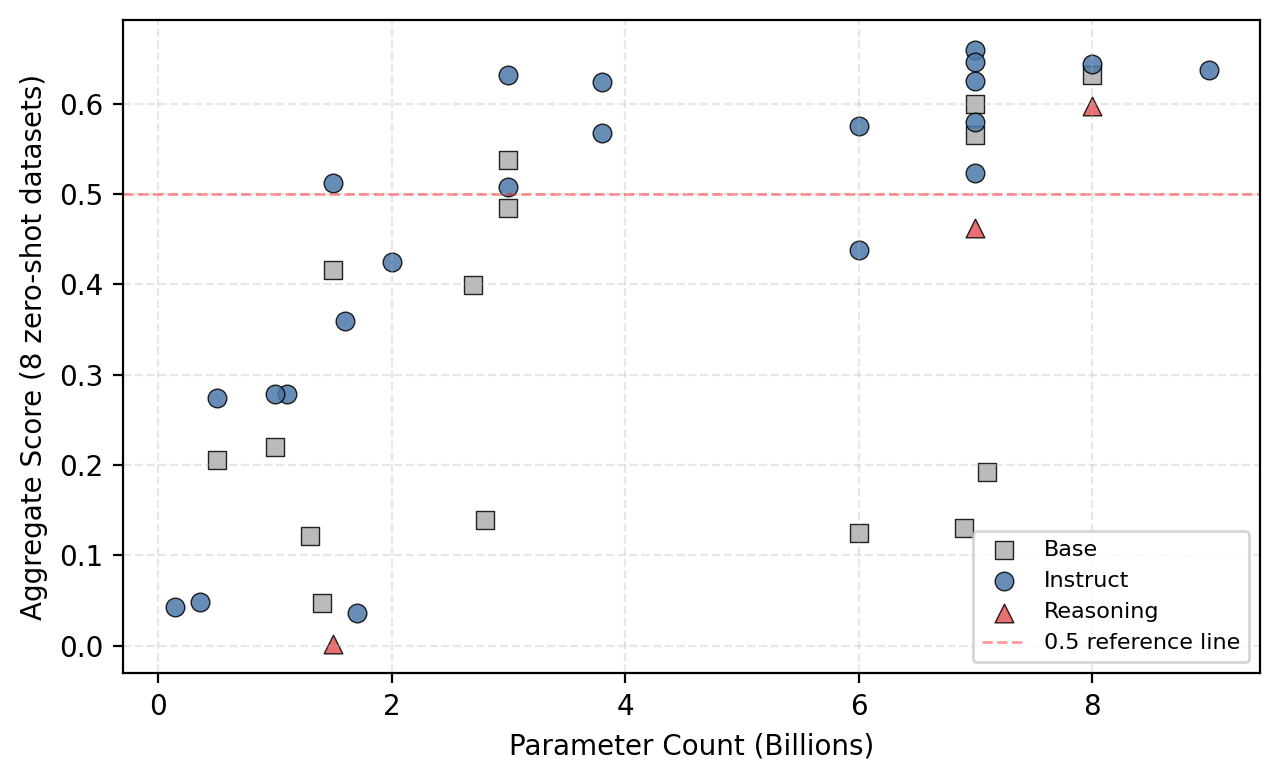}
\caption{Parameter count versus aggregate score for the 40 comparable
models (Qwen2.5-1.5B-Instruct omitted; 8-dataset zero-shot aggregate, ATIS excluded).
Circles = Instruct models, Squares = Base models, Triangles = Reasoning
models. The dashed line is a 0.5 reference line (illustrative, not a
validated deployment criterion). Instruct
models generally outperform base models at the same parameter scale:
all six cleanly comparable pairs improve, no ties, no reversals; see
the main paper's Instruction Tuning section.}
\label{fig:scale}
\end{figure*}

\subsection{Dataset-Group Accuracy Heatmap}
\label{app:heatmap}

\begin{landscape}
\begin{figure}[h]
\centering
\includegraphics[height=0.85\textheight]{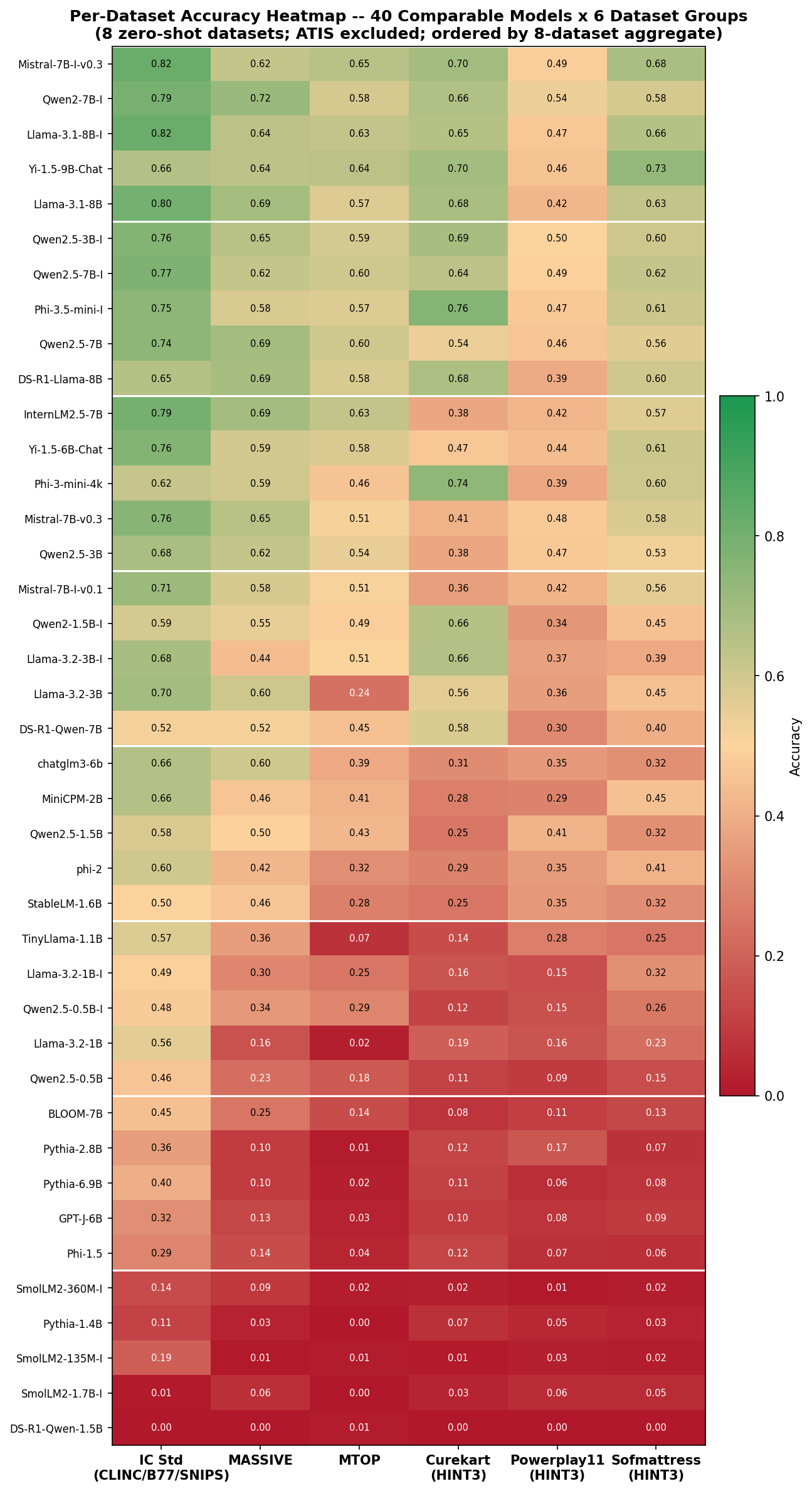}
\caption{Dataset-group accuracy heatmap: 40 comparable models
(Qwen2.5-1.5B-Instruct omitted; rows, ordered by
the 8-dataset zero-shot aggregate score) $\times$ 6 dataset groups (columns; the first column
is the mean of the three zero-shot standard IC benchmarks CLINC150, Banking77,
SNIPS---ATIS is excluded from this composite and from the aggregate
entirely, see the main paper's Overall Model Rankings subsection---so
this is a dataset-group heatmap rather than all eight zero-shot
datasets shown individually; see the main paper's Benchmark Saturation
Analysis for individual per-dataset saturation statistics). Each cell shows
the exact accuracy value. Green = high, Red = low. Curekart and Sofmattress
show the widest top-to-bottom ranges among the individually-shown
columns; Powerplay11 is uniformly the lowest-accuracy column,
consistent with it being the hardest rather than the most
discriminating dataset.}
\label{fig:heatmap_app}
\end{figure}
\end{landscape}

Figure~\ref{fig:heatmap_app} presents the dataset-group accuracy
heatmap for the 40 comparable models across six dataset groups,
ordered by the final eight-dataset zero-shot aggregate. Each cell shows
the exact accuracy value. Green cells indicate high accuracy; red cells
indicate low accuracy. The heatmap is presented in the appendix at full
size for readability.

\subsection{Ranking Confidence Intervals}
\label{app:bootstrap}
\begin{figure*}[h]
\centering
\includegraphics[width=0.85\textwidth]{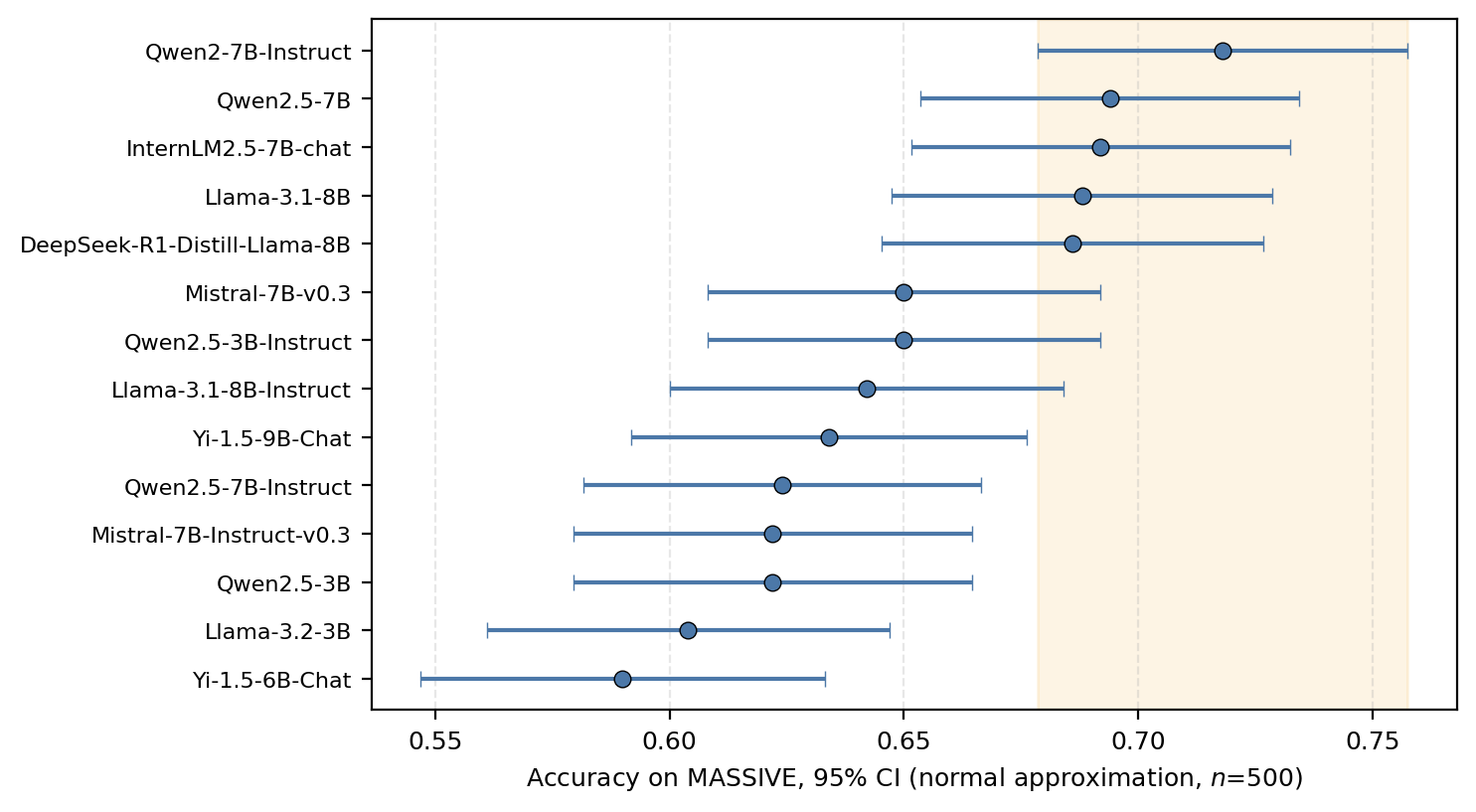}
\caption{MASSIVE accuracy ($n$=500) for the top 14 models, with 95\%
confidence intervals (normal/Wald approximation,
$p\pm1.96\sqrt{p(1-p)/n}$). Point estimates and error bars are shown
rather than bars from zero. The shaded band marks the top model's CI
range; the overlapping intervals illustrate that MASSIVE alone does
not clearly distinguish several leading models (Finding-4).}
\label{fig:bootstrap_app}
\end{figure*}

\subsection{Error Taxonomy on MASSIVE}
\label{app:errors}
\begin{figure*}[h]
\centering
\includegraphics[width=0.7\textwidth]{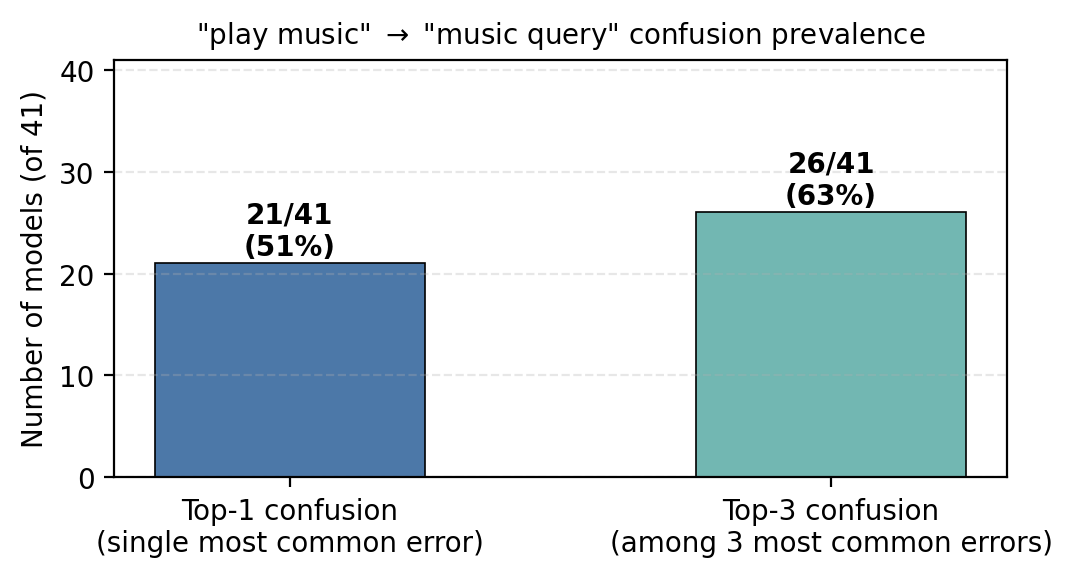}
\caption{Prevalence of the \textit{play music}$\to$\textit{music
query} confusion across all 41 evaluated models on MASSIVE (500
samples): the number of models for which this pair is their single
most common error (left bar) versus among their top-three most
common errors (right bar). This is the dominant, cross-family,
cross-scale confusion pattern discussed in the main paper
(main paper's Systematic Error Patterns subsection); both labels refer to music-related actions, consistent
with a label-scheme ambiguity rather than a model-specific weakness.}
\label{fig:errors_app}
\end{figure*}

\subsection{Deployment Efficiency}
\label{app:efficiency}

\begin{table*}[t]
\centering
\caption{Deployment efficiency for the 40 standard-protocol models
(Qwen2.5-1.5B-Instruct excluded; see Full Model Rankings above). VRAM = peak GPU
memory (GB) during inference. Agg = 8-dataset zero-shot aggregate
(ATIS excluded). Acc/B = aggregate score per billion
parameters. P = Pareto-optimal (Y/N): no other model among these 40
is simultaneously at least as accurate, at least as fast, and at
least as parameter-efficient. Sorted by Acc/B descending.}
\label{tab:efficiency}
\setlength{\tabcolsep}{3pt}
\small
\begin{tabular}{lrrrrrc}
\toprule
\textbf{Model} & \textbf{Params(B)} & \textbf{VRAM} & \textbf{ms/samp} & \textbf{Agg} & \textbf{Acc/B} & \textbf{P} \\
\midrule
Qwen2.5-0.5B-I     & 0.50 &  0.93 &   9.0 & .274 & .548 & Y \\
Qwen2.5-0.5B       & 0.50 &  0.93 &   6.7 & .205 & .410 & Y \\
Qwen2-1.5B-I       & 1.50 &  2.89 &  16.0 & .512 & .341 & Y \\
SmolLM2-135M-I     & 0.14 &  0.26 &   9.7 & .043 & .307 & Y \\
Llama-3.2-1B-I     & 1.00 &  2.32 &  10.4 & .279 & .279 & Y \\
Qwen2.5-1.5B       & 1.50 &  2.91 &  13.6 & .416 & .277 & Y \\
TinyLlama-1.1B     & 1.10 &  2.05 &  11.3 & .279 & .254 & N \\
StableLM-1.6B      & 1.60 &  3.06 &  13.9 & .359 & .224 & N \\
Llama-3.2-1B       & 1.00 &  2.32 &   9.7 & .220 & .220 & N \\
MiniCPM-2B         & 2.00 &  5.10 &  24.4 & .425 & .212 & N \\
Qwen2.5-3B-I       & 3.00 &  5.79 &  20.8 & .632 & .211 & Y \\
Qwen2.5-3B         & 3.00 &  5.79 &  20.6 & .538 & .179 & Y \\
Llama-3.2-3B-I     & 3.00 &  6.02 &  39.6 & .508 & .169 & N \\
Phi-3.5-mini-I     & 3.80 &  7.17 &  40.0 & .624 & .164 & N \\
Llama-3.2-3B       & 3.00 &  6.02 &  21.6 & .485 & .162 & N \\
Phi-3-mini-4k-I    & 3.80 &  7.12 &  31.9 & .568 & .149 & N \\
phi-2              & 2.70 &  5.19 &  28.9 & .399 & .148 & N \\
SmolLM2-360M-I     & 0.36 &  0.67 &  10.6 & .048 & .133 & Y \\
Yi-1.5-6B-Chat     & 6.00 & 11.29 &  50.3 & .575 & .096 & N \\
Mistral-7B-I-v0.3  & 7.00 & 13.51 &  48.5 & .660 & .094 & Y \\
Phi-1.5            & 1.30 &  2.64 &  12.5 & .121 & .093 & N \\
Qwen2-7B-I         & 7.00 & 14.25 &  55.0 & .646 & .092 & N \\
Qwen2.5-7B-I       & 7.00 & 14.25 &  47.7 & .625 & .089 & N \\
Qwen2.5-7B         & 7.00 & 14.25 &  47.7 & .600 & .086 & N \\
InternLM2.5-7B     & 7.00 & 14.43 &  54.5 & .580 & .083 & N \\
Mistral-7B-v0.3    & 7.00 & 13.51 &  58.0 & .565 & .081 & N \\
Llama-3.1-8B-I     & 8.00 & 14.99 &  45.6 & .644 & .081 & Y \\
Llama-3.1-8B       & 8.00 & 14.99 &  49.1 & .632 & .079 & N \\
DS-R1-Llama-8B     & 8.00 & 14.99 &  49.7 & .598 & .075 & N \\
Mistral-7B-I-v0.1  & 7.00 & 13.50 &  57.9 & .523 & .075 & N \\
chatglm3-6b        & 6.00 & 11.66 &  45.9 & .438 & .073 & N \\
Yi-1.5-9B-Chat     & 9.00 & 16.45 &  78.5 & .638 & .071 & N \\
DS-R1-Qwen-7B      & 7.00 & 14.27 &  53.7 & .463 & .066 & N \\
Pythia-2.8B        & 2.80 &  5.19 &  22.4 & .139 & .050 & N \\
Pythia-1.4B        & 1.40 &  2.64 &  12.1 & .047 & .034 & N \\
BLOOM-7B           & 7.10 & 13.17 &  54.4 & .192 & .027 & N \\
SmolLM2-1.7B-I     & 1.70 &  3.19 &  15.4 & .036 & .021 & N \\
GPT-J-6B           & 6.00 & 11.27 &  45.9 & .125 & .021 & N \\
Pythia-6.9B        & 6.90 & 12.77 &  51.2 & .130 & .019 & N \\
DS-R1-Qwen-1.5B    & 1.50 &  3.35 &  15.8 & .002 & .001 & N \\
\bottomrule
\end{tabular}
\end{table*}

\begin{figure*}[h]
\centering
\includegraphics[width=0.75\textwidth]{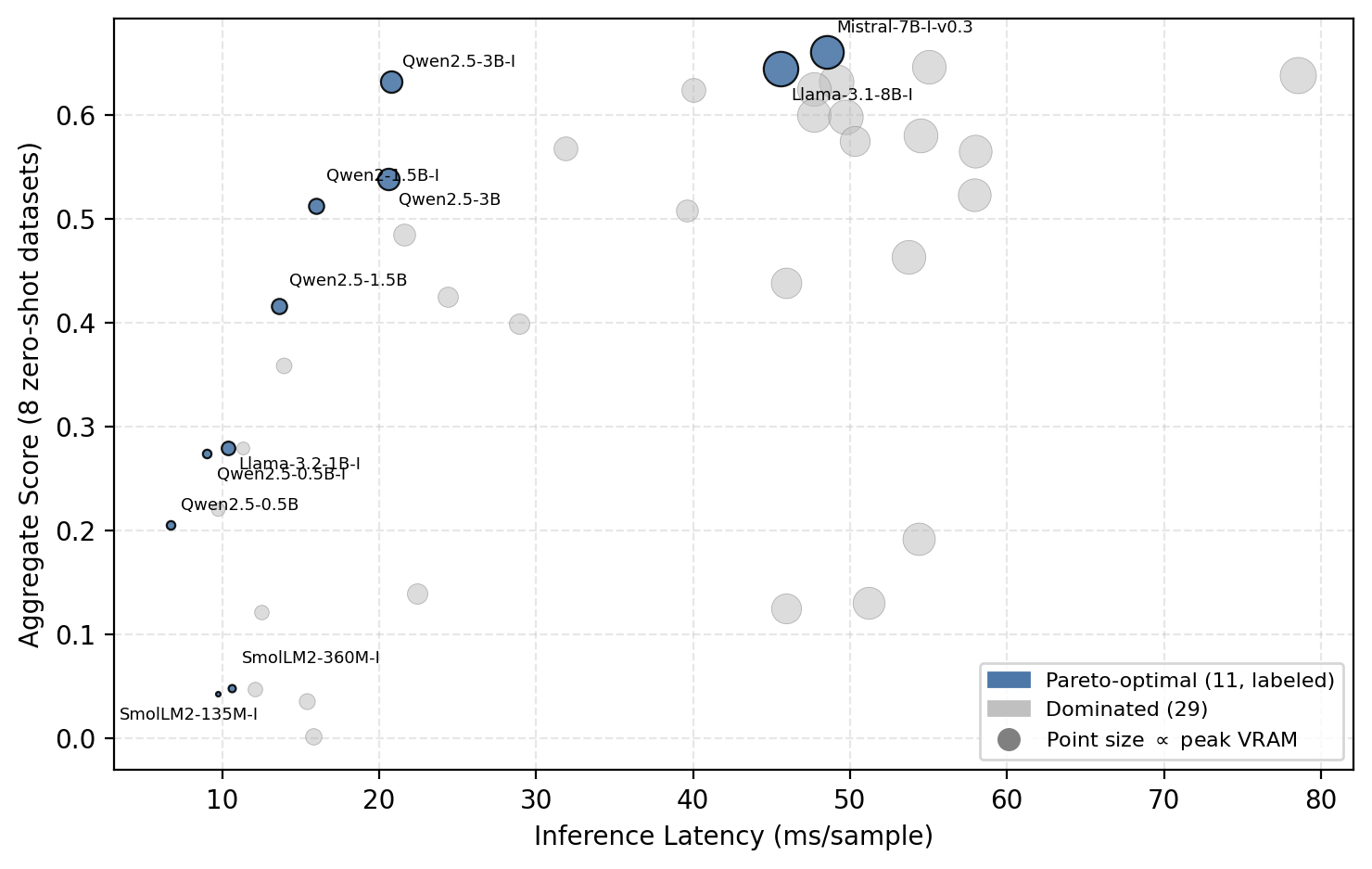}
\caption{Inference latency vs.\ 8-dataset zero-shot aggregate accuracy
for the 40 standard-protocol models (Qwen2.5-1.5B-Instruct excluded).
Point size is proportional to peak GPU memory (VRAM). Blue,
labeled points are the 11 Pareto-optimal models (no other model is
simultaneously at least as accurate, as fast, and as
parameter-efficient); gray points are dominated by at least one
Pareto-optimal model.}
\label{fig:efficiency_app}
\end{figure*}

\subsection{Robustness to Input Perturbations}
\label{app:robustness}
\begin{figure*}[h]
\centering
\includegraphics[width=0.92\textwidth]{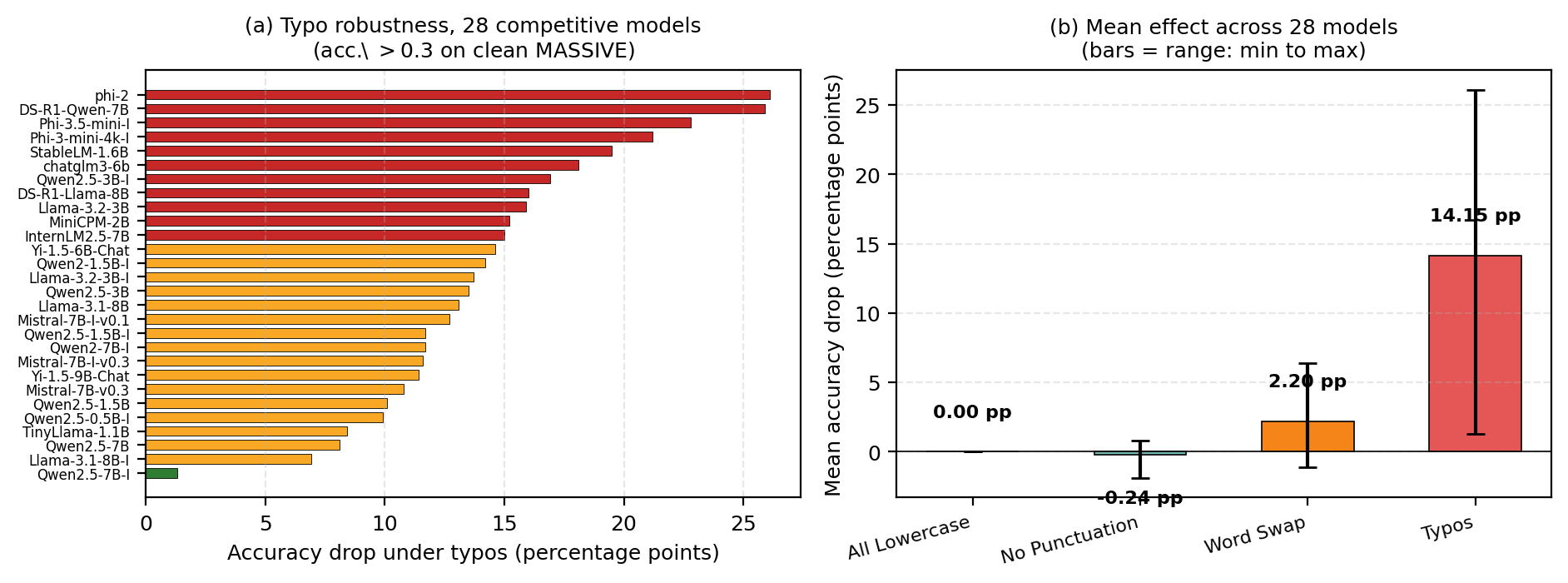}
\caption{Robustness to input perturbations on MASSIVE (500 samples).
(a) Individual typo-induced accuracy drop, in percentage points, for the 28 models with
competitive clean accuracy ($>$0.3); bar color indicates severity
(green $<$5\,pp, amber 5--15\,pp, red $>$15\,pp). (b) Mean accuracy drop, in percentage points,
across the same 28 models for all four perturbation types
(range bars show the full min--max range across models, not a
statistical uncertainty interval); all-lowercase and
punctuation removal have negligible mean effect, word swap has a
small consistent effect, and typos are the dominant and most
variable perturbation.}
\label{fig:robustness_app}
\end{figure*}

\section{Prompt Templates}
\label{app:prompts}

The eight zero-shot prompt templates are reproduced below with label
lists abbreviated for space; instructional wording and formatting are
given verbatim, transcribed directly from the live
\texttt{skills.py} configuration used for the actual evaluation run.
ATIS is described separately because it is an auxiliary five-shot
evaluation.

\paragraph{CLINC150.}
\begin{verbatim}
You are an intent classifier.
Classify the user query into
exactly one of these labels:
[restaurant_reviews, ..., 150 total]

Reply with the label only, nothing
else. Use underscores not spaces.
Query: {input}
Label:
\end{verbatim}

\paragraph{Banking77.}
\begin{verbatim}
You are an intent classifier for
a banking assistant.
Classify the user query into
exactly one of these labels:
[activate_my_card, ..., 77 total]

Reply with the label only, nothing
else. Use underscores not spaces.
Query: {input}
Label:
\end{verbatim}

\paragraph{ATIS.}
ATIS's prompt template embeds five labeled examples before the
target query, unlike the other eight datasets; ATIS results in this
paper therefore reflect 5-shot in-context performance rather than
zero-shot and are excluded from all aggregate scores, rankings, and
comparisons (see the main paper's Results section).

\paragraph{SNIPS.}
\begin{verbatim}
You are an intent classifier.
Classify the user utterance into
exactly one of these intents:
[addtoplaylist, ..., 7 total]

Reply with the intent label only,
nothing else. Use lowercase no
spaces.
Utterance: {input}
Intent:
\end{verbatim}
Note SNIPS uses \texttt{Utterance:}/\texttt{Intent:}, not
\texttt{Query:}/\texttt{Label:} as most other datasets do.

\paragraph{MASSIVE.}
\begin{verbatim}
You are an intent classifier for
a voice assistant.
Classify the user query into
exactly one of these labels:
[alarm_query, ..., 60 total]

Reply with the label only, nothing
else. Use spaces not underscores.
Query: {input}
Label:
\end{verbatim}

\paragraph{MTOP.}
\begin{verbatim}
You are an intent classifier for
a voice and messaging assistant.
Classify the user query into
exactly one of these labels:
[add time timer, ..., 117 total]

Reply with the label only, nothing
else.
Query: {input}
Label:
\end{verbatim}
Note MTOP omits the spaces-vs-underscores instruction present in
several other templates (its labels are already space-separated in
\texttt{skills.py}).

\paragraph{HINT3 (Curekart, Powerplay11, Sofmattress).}
\begin{verbatim}
You are an intent classifier for
a [health supplement e-commerce /
online fantasy gaming platform /
mattress e-commerce] chatbot.
Classify the user query into
exactly one of these labels:
[CALL_CENTER, ..., 20/53/20 total]

Reply with the label only, nothing
else. Use uppercase.
Query: {input}
Label:
\end{verbatim}
All three HINT3 datasets share this identical structure, differing
only in the domain phrase and label list.

\paragraph{Answer-parsing logic.} Confirmed directly from
\texttt{evaluator.py}: generated text is split on the literal string
``query:'' (case-insensitive), and the portion before the first such
occurrence is taken as the answer, then stripped and
lowercased/underscore-normalized as needed per dataset. This is a
no-op for the common case where the model does not echo the word
``query'' in its own output; it is a safeguard against models that
echo the input prompt structure before answering. Invalid outputs
were scored as incorrect; per-model invalid-output rates are not
reported separately.

\section{Statistical Significance}
\label{app:stats}

We report the ten verified pairwise McNemar's test comparisons (on
matched predictions, $n$=500, MASSIVE) among the top-five models by
MASSIVE accuracy, reported in the main paper's Finding-4
($p$=0.1422 to $p$=0.3384, all non-significant); a full pairwise
matrix across all models is outside the scope of this analysis.
Extending pairwise significance testing to the full
model set, with an appropriate multiple-comparison correction given
the large number of pairs, is left to future work.

\section{Implementation Details}
\label{app:impl}

\paragraph{Hardware.} All experiments were conducted on the PSC
Bridges-2 HPC cluster using NVIDIA V100 GPUs with 32 GB of memory.
Each model--dataset inference job used a single GPU and was executed
in FP16 using vLLM 0.6.3. The complete evaluation required
approximately 320 V100 GPU-hours across 41 models and nine datasets.
Mean per-sample inference latency and peak GPU-memory usage for the
40 standard-protocol models are reported in Supplementary Table 4.

\paragraph{Software.} vLLM v0.6.3, Python 3.10, float16 precision
(bfloat16 unavailable on V100), XFormers attention backend. All
generation is free-text (greedy decoding, max 20 tokens, stop tokens
\texttt{["\textbackslash n", "</s>", "<|endoftext|>"]}); no constrained
decoding is applied on any dataset. Random seed = 0 throughout.

\paragraph{Data Sampling.} First 500 examples from each test split
selected deterministically by index. HINT3: complete test sets (459,
309, and 253 examples for Curekart, Powerplay11, and Sofmattress).
Per-example analyses use the same 500-example subsets.

\paragraph{Reproducibility.} All predictions, logprobabilities,
SLURM job scripts, and evaluation code will be released upon
acceptance at the project repository.

\end{document}